\documentclass[]{ecai} 

\usepackage{array}
\usepackage{backnaur}
\usepackage{microtype}
\usepackage{algpseudocode}
\usepackage{subfigure}
\usepackage{hyperref}
\usepackage{algorithm}
\usepackage{latexsym}
\usepackage{amssymb}
\usepackage{mathtools}
\usepackage{amsmath}
\usepackage{amsthm}
\usepackage{booktabs}
\usepackage{enumitem}
\usepackage{graphicx}
\usepackage{color}
\usepackage{multirow}
\usepackage[numbers]{natbib}
\usepackage{tikz}
\usepackage{tablefootnote}
\usetikzlibrary{shapes.geometric, arrows, positioning}

\makeatletter
\renewcommand{\ALG@name}{Example}
\makeatother

\usepackage{listings,caption,syntax}
\lstdefinestyle{promptstyle}{
  basicstyle=\ttfamily\small,
  breaklines=true,
  breakindent=0pt,         
  breakautoindent=false,   
  columns=fullflexible,
  frame=single,
  backgroundcolor=\color{gray!10},
  literate={\\n}{{\unskip\newline\noindent}}2
}

\usepackage{multicol}

\usepackage[disable]{todonotes}

\usepackage{algorithm}
\usepackage{algorithmicx}
\usepackage{algpseudocode}

\newcommand{\BibTeX}{B\kern-.05em{\sc i\kern-.025em b}\kern-.08em\TeX}

\definecolor{lightgray}{gray}{0.9}

\begin{document}


\begin{frontmatter}


\paperid{4057} 


\title{Grammar-Guided Evolutionary Search for\newline Discrete Prompt Optimisation}


\author[A,B]{\fnms{Muzhaffar}~\snm{Hazman}\footnote{Work conducted as interns at Huawei Technologies, Ireland Research Centre.}\thanks{Corresponding Author. Email: m.hazman1@universityofgalway.ie.}}
\author[A,C]{\fnms{Minh-Khoi}~\snm{Pham}\footnotemark}
\author[A,D]{\fnms{Shweta}~\snm{Soundararajan}\footnotemark} \author[A]{\fnms{Goncalo}~\snm{Mordido}} \author[A]{\fnms{Leonardo}~\snm{Custode}} \author[A]{\fnms{David}~\snm{Lynch}} \author[A]{\fnms{Giorgio}~\snm{Cruciata}} \author[A]{\fnms{Yucheng}~\snm{Shi}} \author[A]{\fnms{Hongmeng}~\snm{Song}} \author[A]{\fnms{Wang}~\snm{Chao}} \author[A]{\fnms{Pan}~\snm{Yue}} \author[A]{\fnms{Aleksandar}~\snm{Milenovic}} \author[A]{\fnms{Alexandros}~\snm{Agapitos}} 

\address[A]{Huawei Technologies, Ireland Research Center, Ireland}
\address[B]{University of Galway, Ireland}
\address[C]{Dublin City University, Ireland}
\address[D]{Technological University Dublin, Ireland}


\begin{abstract}
Prompt engineering has proven to be a crucial step in leveraging pretrained large language models (LLMs) in solving various real-world tasks. Numerous solutions have been proposed that seek to automate prompt engineering by using the model itself to edit prompts. However, the majority of state-of-the-art approaches are evaluated on tasks that require minimal prompt templates and on very large and highly capable LLMs. In contrast, solving complex tasks that require detailed information to be included in the prompt increases the amount of text that needs to be optimised. Furthermore, smaller models have been shown to be more sensitive to prompt design. To address these challenges, we propose an evolutionary search approach to automated discrete prompt optimisation consisting of two phases. In the first phase, grammar-guided genetic programming is invoked to synthesise prompt-creating programmes by searching the space of programmes populated by function compositions of syntactic, dictionary-based and LLM-based prompt-editing functions. In the second phase, local search is applied to explore the neighbourhoods of best-performing programmes in an attempt to further fine-tune their performance. Our approach outperforms three state-of-the-art prompt optimisation approaches, PromptWizard, OPRO, and RL-Prompt, on three relatively small general-purpose LLMs in four domain-specific challenging tasks. We also illustrate several examples where these benchmark methods suffer relatively severe performance degradation, while our approach improves performance in almost all task-model combinations, only incurring minimal degradation when it does not.
\end{abstract}

\end{frontmatter}


\section{Introduction}
\label{sec:introduction}

The advent of pretrained large language models (LLMs) as general-purpose Natural Language Inferencing models that are performant in a wide range of tasks has prompted both academic and industrial use of these models to tackle challenging tasks. Deploying such models typically requires prompt engineering that customises the behaviour of the model by manually defining task parameters within prompt templates \cite{giray2023promptengineeringguide}. Manual trial-and-error prompt engineering is feasible for many general natural language tasks, mitigating the need for parameter finetuning. However, complex or niche tasks often require longer prompts to which a model's performance can be highly sensitive. Optimising such prompt templates is not trivial, as the search space is large and evaluating candidate solutions may be expensive
\cite{sahoo2024surveypromptengineering,sclar2024spuriousformatting}.

In pursuit of better task performance of pretrained LLMs, the high computational costs associated with finetuning have inspired a plethora of \textit{black-box} prompt engineering techniques that treat the internal weights of any given LLMs as untrainable and unaccessible \cite{sahoo2024surveypromptengineering}. Recent work has shown that the performance of LLMs across various tasks can be further improved through \textit{prompt optimisation} -- editing and adjusting prompt templates with the goal of improving task performance. Algorithms for automatic prompt optimisation fall into two main categories; where optimisation is performed over prompt embeddings, or by directly editing the natural language description the seed prompt. These classes of methods are known as \textit{ soft prompt tuning} and \textit{ automatic discrete prompt optimisation} (DPO), respectively \cite{sahoo2024surveypromptengineering}.

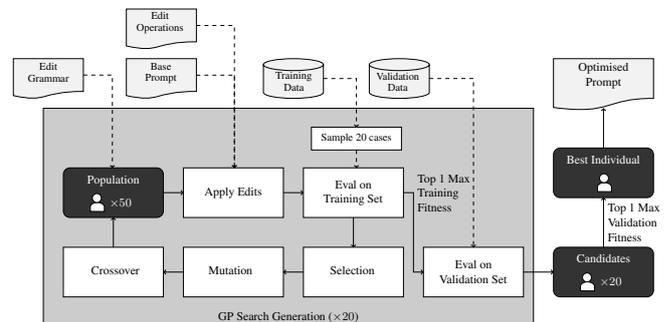
\begin{figure}[b]
    \centering
        \begin{minipage}[b]{\linewidth}
        \centering
        \resizebox{\linewidth}{!}{




\begin{tikzpicture}[x=0.75pt,y=0.75pt,yscale=-1,xscale=1]

\draw[fill=black!20]    (50,110) -- (540,110) -- (540,330) -- (50,330) -- cycle ;
\node[anchor=south] at (295,330) {GP Search Generation ($\times20$)};
\draw[fill=black!5]   (20,60) -- (90,60) -- (90,93) .. controls (46.25,93) and (55,104.9) .. (20,97.2) -- cycle ;
\draw[dashed,->] (90, 76.5) -| (119,170);
\draw[rounded corners=5pt,fill=black!80]   (70,170) -- (170,170) -- (170,220) -- (70,220) -- cycle ;
\draw[->] (170,195) -- (190,195);
\draw[fill=white]   (96.93,213.9) -- (96.96,209.38) .. controls (96.98,206.88) and (100.14,204.87) .. (104.03,204.9) .. controls (107.91,204.93) and (111.05,206.97) .. (111.03,209.47) -- (111,214) -- cycle ;
\draw[fill=white]   (99,200.4) .. controls (99,197.91) and (101.01,195.9) .. (103.5,195.9) .. controls (105.99,195.9) and (108,197.91) .. (108,200.4) .. controls (108,202.88) and (105.99,204.9) .. (103.5,204.9) .. controls (101.01,204.9) and (99,202.88) .. (99,200.4) -- cycle ;
\draw[fill=white!10] (190,170) -- (290,170) -- (290,220) -- (190,220) -- cycle;

\draw (240,195) node [anchor=center] {Apply Edits};

\draw[->] (290,195) -- (310,195);
\draw[fill=white!10] (310,170) -- (410,170) -- (410,220) -- (310,220) -- cycle;

\draw (360,195) node [anchor=center] {\begin{minipage}{57.34pt}\centering
Eval on\\Training Set
\end{minipage}};

\draw[->] (360,220) -- (360,250);
\draw[->] (410,195) -- (420,195) node[right,align=left] {Top 1 Max\\Training\\Fitness} -- (420,275) -- (430,275)  ;
\draw[fill=white!10] (430,250) -- (530,250) -- (530,300) -- (430,300) -- cycle;
\draw[->] (530,275) -- (560,275);

\draw (480,275) node [anchor=center] {\begin{minipage}{64.91pt}\centering
Eval on\\Validation Set
\end{minipage}};

\draw[fill=black!5]   (133,60) -- (203,60) -- (203,93) .. controls (159.25,93) and (168,104.9) .. (133,97.2) -- cycle; 

  \draw[dashed,->] (203,76.6) -| (241,170); 
\draw[fill=black!5]   (133,10) -- (203,10) -- (203,43) .. controls (159.25,43) and (168,54.9) .. (133,47.2) -- cycle ;
  \draw[dashed,->] (203,26.6) -| (241,170); 
\draw[fill=black!5]   (330,66) -- (330,94) .. controls (330,97.31) and (316.57,100) .. (300,100) .. controls (283.43,100) and (270,97.31) .. (270,94) -- (270,66) .. controls (270,62.69) and (283.43,60) .. (300,60) .. controls (316.57,60) and (330,62.69) .. (330,66) .. controls (330,69.31) and (316.57,72) .. (300,72) .. controls (283.43,72) and (270,69.31) .. (270,66) ;
\draw[dashed,->] (330,80) -| (363,129); 

\draw[fill=black!5] (435.86,66) -- (435.86,94) .. controls (435.86,97.31) and (422.43,100) .. (405.86,100) .. 
  controls (389.29,100) and (375.86,97.31) .. (375.86,94) -- (375.86,66) .. controls (375.86,62.69) and (389.29,60) .. (405.86,60) .. 
  controls (422.43,60) and (435.86,62.69) .. (435.86,66) .. controls (435.86,69.31) and (422.43,72) .. (405.86,72) .. 
  controls (389.29,72) and (375.86,69.31) .. (375.86,66);
\draw (405.86,81) node [anchor=center][inner sep=0.75pt] [font=\footnotesize] [align=left] {
  \begin{minipage}[lt]{31.9pt}\setlength\topsep{0pt}
    \begin{center}
      Validation\\Data
    \end{center}
  \end{minipage}
};

\draw [dashed,->](435.86,80)-| (480,250);

\draw[fill=white!10]    (310,250) -- (410,250) -- (410,300) -- (310,300) -- cycle ;
\draw[fill=white!10]    (190,250) -- (290,250) -- (290,300) -- (190,300) -- cycle ;
\draw[->] (310,275) -- (290,275); 
\draw[->] (190,275) -- (170,275); 

\draw[fill=white!10]   (70,250) -- (170,250) -- (170,300) -- (70,300) -- cycle ;
\draw[->] (120,250) -- (120,220);
\draw[rounded corners=5pt,fill=black!80]   (560,150) -- (660,150) -- (660,200) -- (560,200) -- cycle ;
\draw[->] (610,250) -- (610,200) node[midway, right,align=left] {Top 1 Max\\Validation\\Fitness};
\draw[fill=white!10]     (318,129) -- (408,129) -- (408,152) -- (318,152) -- cycle  ;
\draw (363,140.5) node [anchor=center][inner sep=0.75pt]  [font=\small] [align=center] {{\footnotesize Sample 20 cases}};
\draw[dashed,->] (363,152) -| (363,170); 

\draw[rounded corners=5pt,fill=black!80]   (560,250) -- (660,250) -- (660,300) -- (560,300) -- cycle ;

\draw[fill=white]   (603.93,193.9) -- (603.96,189.38) .. controls (603.98,186.88) and (607.14,184.87) .. (611.03,184.9) .. controls (614.91,184.93) and (618.05,186.97) .. (618.03,189.47) -- (618,194) -- cycle ;

\draw[fill=white]   (589,280.4) .. controls (589,277.91) and (591.01,275.9) .. (593.5,275.9) .. controls (595.99,275.9) and (598,277.91) .. (598,280.4) .. controls (598,282.88) and (595.99,284.9) .. (593.5,284.9) .. controls (591.01,284.9) and (589,282.88) .. (589,280.4) -- cycle ;

\draw[fill=white]   (586.93,293.9) -- (586.96,289.38) .. controls (586.98,286.88) and (590.14,284.87) .. (594.03,284.9) .. controls (597.91,284.93) and (601.05,286.97) .. (601.03,289.47) -- (601,294) -- cycle ;

\draw[fill=white]   (606,180.4) .. controls (606,177.91) and (608.01,175.9) .. (610.5,175.9) .. controls (612.99,175.9) and (615,177.91) .. (615,180.4) .. controls (615,182.88) and (612.99,184.9) .. (610.5,184.9) .. controls (608.01,184.9) and (606,182.88) .. (606,180.4) -- cycle ;

\draw (29,63) node [anchor=north west][inner sep=0.75pt]  [font=\footnotesize] [align=left] {\begin{minipage}[lt]{37.18pt}\setlength\topsep{0pt}
\begin{center}
Edit\\Grammar
\end{center}

\end{minipage}};
\draw (120,176) node [anchor=north][inner sep=0.75pt]   [align=center] {\textcolor{white}{Population}\\};
\draw (115,200) node [anchor=north west][inner sep=0.75pt]   [align=left] {{\fontfamily{pcr}\selectfont {\small \textcolor{white}{$\times50$}}}};

\draw (328,268) node [anchor=north west][inner sep=0.75pt]   [align=left] {\begin{minipage}[lt]{44.68pt}\setlength\topsep{0pt}
\begin{center}
Selection
\end{center}

\end{minipage}};
\draw (208,268) node [anchor=north west][inner sep=0.75pt]   [align=left] {\begin{minipage}[lt]{41.85pt}\setlength\topsep{0pt}
\begin{center}
Mutation
\end{center}

\end{minipage}};
\draw (88,268) node [anchor=north west][inner sep=0.75pt]   [align=left] {\begin{minipage}[lt]{49.2pt}\setlength\topsep{0pt}
\begin{center}
Crossover
\end{center}

\end{minipage}};
\draw (610,156) node [anchor=north][inner sep=0.75pt]   [align=center] {\textcolor{white}{Best Individual}};
\draw (605,285) node [anchor=west][inner sep=0.75pt] [align=left] {{\fontfamily{pcr}\selectfont {\small \textcolor{white}{$\times20$}}}};

\draw (610,256) node [anchor=north][inner sep=0.75pt]   [align=center] {\textcolor{white}{Candidates}};
\draw (147,63) node [anchor=north west][inner sep=0.75pt]  [font=\footnotesize] [align=left] {\begin{minipage}[lt]{29.03pt}\setlength\topsep{0pt}
\begin{center}
Base\\Prompt
\end{center}
\end{minipage}};
\draw (137,12) node [anchor=north west][inner sep=0.75pt]  [font=\footnotesize] [align=left] {\begin{minipage}[lt]{42.64pt}\setlength\topsep{0pt}
\begin{center}
Edit\\Operations
\end{center}

\end{minipage}};
\draw (278,72) node [anchor=north west][inner sep=0.75pt]  [font=\footnotesize] [align=left] {\begin{minipage}[lt]{31.9pt}\setlength\topsep{0pt}
\begin{center}
Training\\Data
\end{center}

\end{minipage}};

\draw[fill=black!5]   (560,60) -- (660,60) -- (660,100) .. controls (610,100) and (620,116.9) .. (560,109.2) -- cycle;

\draw (610,63) node [anchor=north, align=center, inner sep=0.75pt] { Optimised\\ Prompt};
\draw[->] (610,150) -- (610,110);

\end{tikzpicture}

        }

        \caption{\small Our Grammar-Guided Genetic Programming for Discrete Prompt Optimisation (G3P DPO) where each individual represents an edited version of a prompt. The search iteratively produces and evaluates individuals, returning the fittest as the task-optimised prompt.}
        \label{fig:g3p}        
    \end{minipage}

\end{figure}

Previous works in DPO are numerous and include approaches that apply search strategies in broad (potentially infinite) search spaces \cite{deng2022rlprompt}, constrained optimisation to parts of a prompt template \cite{lu2022fantasticallyorderedprompt}, limited text editing to syntactic operations \cite{zhang2022tempera}, or applied edits at a single preset chunking level \cite{prasad2022grips}. In particular, most of the work in this area evaluated DPO approaches on tasks that require a short prompt template \cite{hsieh2023automatic}. For instance, \citeauthor{deng2022rlprompt} limited their search to only 5 tokens \cite{deng2022rlprompt}, and \citeauthor{yang2024opro} limited prompt templates to 500 characters \cite{yang2024opro}. Although numerous DPO methods have been evaluated on challenging tasks \cite{ramnath2025systematicsurveyautomaticprompt}, many of these tasks are posed in a way that requires prompt templates of minimal length. 

Furthermore, DPO methods that employ LLM-based editing typically use very large models to edit and evaluate their prompts \cite{agarwal2024promptwizard,fernando2023apopromptbreeder,hsieh2023automatic,yang2024opro}. This is likely due to their reliance on the large model's ability to identify weaknesses within a prompt and the edits needed to improve it. PromptWizard, a particularly effective recent DPO method, was shown to be reliable in many tasks when paired with GPT-4, GPT3.5, or Llama 70B, but resulted in significant performance degradation when applied to Llama3 8B \cite{agarwal2024promptwizard}.

Although DPO methods that employ LLM-based editing have demonstrated effectiveness on large-scale models, their application to smaller LLMs remains a significant gap; see Tab. \ref{tab:modelsizes}. Concurrently, DPO techniques without LLM-based editing seldom address tasks that require lengthy and intricate prompt templates. Adapting smaller LLMs to such tasks is attractive due to their reduced compute requirements and accelerated inference, but these models exhibit a pronounced sensitivity to prompt formulation \cite{zhuo2024prosa}. Consequently, without suitable optimisation methods, performance degradation is likely when opting for smaller models. Bridging this gap necessitates DPO strategies that are designed and evaluated in resource-constrained scenarios.

In this work, we tackle the optimisation of \textbf{longer prompts} for technically challenging tasks using \textbf{smaller general purpose LLMs}. Our method formulates the prompt optimisation problem as an evolutionary programme synthesis problem that is solved by means of Grammar-guided Genetic Programming~\cite{whigham1995grammatically} (G3P); see Fig. \ref{fig:g3p}. The search space is populated by compositions of prompt-editing functions consisting of semantic, syntactic, list manipulation, and LLM-based edit operations; see Tab. \ref{tab:operations}. Local search is applied to the best individual found with G3P to further optimise the parameters of its edit operations; see Fig. \ref{fig:local}. The neighbourhood is created as a set of point mutations for each of the parameters of the best performing individual, and the resulting neighbours are screened using a surrogate model of prompt performance. A small number of individuals that maximise exploration or exploitation are selected for real (and expensive LLM-based) fitness evaluation, and the ultimate individual is selected based on validation score. Experiments demonstrate that the proposed system improves the performance of small general-purpose models in complex tasks from various technical domains; while three notable benchmark approaches -- two that use LLM-based edits and one that uses token search -- exhibited a greater propensity for performance degradation when given the same base prompts, tasks, and models.


\section{Related Work}\label{sec:background}

\subsection{LLM Performance Sensitivity to Prompt Design}
LLMs are highly sensitive to prompt design. Significant changes in performance are induced from minor changes to the prompt, such as semantically identical paraphrasing; adding ostensibly spurious changes in text format (e.g., casing, separators, number of spaces) \cite{sclar2024spuriousformatting}, as well as the order of answer choices \cite{zheng2024multiplechoice}, and In-Context Learning (ICL) demonstrations \cite{lu2022fantasticallyorderedprompt}. In particular, this brittleness is especially exacerbated in smaller models as well as complex and domain-specific tasks \cite{zhuo2024prosa}, rendering manual prompt engineering unreliable and unscalable. This sensitivity has also led to methods that have been found to reliably improve task performance, including ICL\cite{brown2020fewshotlearners}, Chain-of-Thought (CoT) prompting, Few-Shot CoT, and Zero-Shot CoT \cite{wei2023chainofthought}. This suite of methods that operate on the LLM context have motivated prompt engineering as a viable approach for improving LLM performance without requiring costly finetuning.

Given the complexity of designing effective prompts that must consider different model capabilities and task requirements, numerous automated prompt optimisation approaches have been proposed. Our approach is an instance of Automated Discrete Prompt Optimisation (DPO), where prompts are treated as a set of discrete tokens; as opposed to Soft Prompt Tuning \cite{sahoo2024surveypromptengineering} which performs optimisation within the LLM's embedding space. DPO approaches generally comprise a set of edit operations, an optimisation algorithm, and a reward signal to guide the edits.

\subsection{LLM-based Editing in DPO}
Several recent works leveraged LLMs to rewrite prompts using self-reflection or few-shot feedback loops. For example, \textsc{ProTeGi} \cite{pryzant2023apoprotegi} instruments an LLM to generate a reflection expressed in natural language based on task errors incurred by a base prompt, similar to methods like \textsc{REFINER} \cite{paul2023refiner}
and PromptAgent \cite{wang2023promptagent}– and using this summary to edit the prompt. OPRO \cite{yang2024opro} and PromptWizard \cite{agarwal2024promptwizard} adopt a similar strategy, but these methods also incorporated ICL demonstrations during the rewriting phase. Building on this, PACE \cite{dong2023pace} frames prompt editing as a policy learning problem, using an actor-critic reinforcement learning setup to iteratively refine discrete prompts based on LLM feedback. 

However, many of these methods rely heavily on frequent LLM calls, making them expensive, less interpretable, and difficult to scale in practice. Although these have been shown to be reliably effective, they often assume access to large proprietary models, limiting their applicability in resource-constrained settings\cite{agarwal2024promptwizard,hsieh2023automatic}. Crucially, the majority of existing DPO approaches, particularly those that rely on LLM-based edits, rely on very large models. Table~\ref{tab:modelsizes} summarises the LLMs used in several prior DPO works, illustrating their reliance on large and often proprietary LLMs.

\subsection{Search-based DPO}

\begin{table}[t]
    \centering
        \caption{LLMs and their sizes as utilised and evaluated in prior DPO works, categorised into: LLM-based Editing (\textbf{LLM-Edit}), LLM-based Evolutionary Analogue (\textbf{LLM-Evo}), and Syntatic/Token/Edit Search (\textbf{STES}). \textbf{*} indicates approximated sizes.}
    \label{tab:modelsizes}
    \begin{tabular}{l >{\raggedright\arraybackslash}m{0.05\columnwidth} >{\raggedright\arraybackslash}m{0.27\columnwidth} >{\raggedright\arraybackslash}m{0.15\columnwidth} >{\raggedright\arraybackslash}m{0.13\columnwidth} }

    \toprule
         \textbf{Model} & \textbf{Size} & \textbf{LLM-Edit} & \textbf{LLM-Evo} & \textbf{STES}\\ 
         \toprule

         GPT-4 & 1.8T* &  \cite{agarwal2024promptwizard,dong2023pace, pryzant2023apoprotegi,wang2023promptagent,yang2024opro} &  \cite{cui2024phaseevo} & \cite{ schnabel2024sammo}
         \\

          PaLM2 & 340B  &  \cite{wang2023promptagent,yang2024opro} & \cite{fernando2023apopromptbreeder} & \cite{kong2024apoprewrite} \\
          
         GPT-3.5 & 175B* &  \cite{agarwal2024promptwizard,dong2023pace, paul2023refiner,pryzant2023apoprotegi,wang2023promptagent,yang2024opro} &  \cite{cui2024phaseevo,fernando2023apopromptbreeder, guo2024apoevoprompt}
         & \cite{li2023guiding,schnabel2024sammo} \\

         GPT-3 & 175B* &  
            
         \cite{pryzant2023apoprotegi} & \cite{pan2024apoplum}& \cite{prasad2022grips}
         \\

         Llama2 & 70B &  \cite{agarwal2024promptwizard}  & &\cite{schnabel2024sammo}
        \\

        Mixtral & 45B & &&  \cite{schnabel2024sammo} \\

          Llama2 & 13B & &  \cite{li2023apospell} \\
          
    Nemo & 12B & & & \cite{zhang2024sprig}\\
        
       T5-XXL & 11B &  &  \cite{xu2022apogps}\\

         Llama3.1 & 8B & & &  \cite{zhang2024sprig} \\

         Llama2 & 7B & & \cite{li2023apospell} \\

         BLOOMZ & 7B & &  \cite{li2023apospell} \\

         Qwen2.5 & 7B & &&  \cite{zhang2024sprig} \\

         Alpaca & 7B & & \cite{guo2024apoevoprompt} \\
         GPT-2 XL & 1.5B & & \cite{xu2022apogps} &  \cite{deng2022rlprompt,prasad2022grips} \\

         \bottomrule

    \end{tabular}

\end{table}

A common approach to DPO is to search for a sequence of edit operations that, when applied to a manually crafted prompt, creates an improved prompt that boosts the LLM's performance on a given task. This class of edit-search DPO methods are practical and lightweight, since each edit operation is performed at the token level, and typically involve minor rewrites and rule-based edits. The underlying search strategies include beam search, reinforcement learning (RL), and genetic search, each offering distinct benefits and drawbacks. 
For instance, in \textsc{Tempera} \cite{zhang2022tempera} the authors frame DPO as an RL problem, with the action space consisting of edits to chunked phrases of the prompt, sequence ordering of ICL demonstrations, and synonymising classification labels. \textsc{GrIPS} \cite{prasad2022grips} uses a similar set of operations but replaces the RL-based optimiser with beam search, while \textsc{SPRIG} \cite{zhang2024sprig} extends this by optimising system prompts using GrIPS' editing operations. \textsc{SAMMO} \cite{schnabel2024sammo} employs enumerative and iterative search algorithms, combining structural, syntactic, and LLM-based edits for a wider variety of actions. The above methods do not chunk the prompt prior to editing or use a single level of chunking to do so; to the best of our knowledge existing methods do not dynamically assigns different chunking levels for each operation.

Search has also been applied to propose individual tokens that make up an optimised prompt. In RL-Prompt \cite{deng2022rlprompt}, the RL agent selects the next token at each step, training a policy model to predict the next token based on a task-specific reward. It uses masked language modelling for text classification by selecting high-probability classification label synonyms predicted by RL policy. PReWrite \cite{kong2024apoprewrite} shares this formulation, but fine-tunes a dedicated LLM for prompt rewriting, while DSP \cite{li2023guiding} instead trains a small policy model to generate directional stimulus prompts guiding the LLM output.

\subsection{Evolutionary Search in DPO}

Evolutionary algorithms are powerful optimisation techniques inspired by natural selection, using processes of mutation, crossover, and selection to efficiently explore complex discrete search spaces \cite{back1997handbook}. Current evolutionary DPO approaches invoke smaller models for LLM-based editing, wherein  evolutionary operators are emulated through prompting strategies. Here, the baseline prompt is regarded as the genome constituting an individual. Updated individuals (prompts) are generated by performing LLM-based edits on individuals (mutation) selected from the population at the previous generation, and by combining two individual's geneomes (via crossover) to create a new individual. Variations of this formulation have been explored \cite{cui2024phaseevo,fernando2023apopromptbreeder,guo2024apoevoprompt,hsieh2023automatic,li2023apospell,xu2022apogps}. However, we are aware of only one DPO approach that does not replace evolutionary reproduction operators with LLM-based analogues. PLUM \cite{pan2024apoplum} tested several metaheuristic algorithms, where instead of applying LLM edits to produce new prompts, mutation and crossover are applied to phrases in the prompt.


Two major challenges have hindered the widespread adoption of evolutionary DPO. Firstly, population-based search using frontier LLMs is computationally infeasible. Secondly, evolutionary operators, where a smaller LLM performs mutation and crossover over the entire prompt, tend to be destructive and often produce unviable prompts \cite{li2023apospell} or degraded performance on certain tasks \cite{guo2024apoevoprompt}. The prospect of leveraging smaller LMs within an evolutionary search procedure is intriguing, since the search becomes computationally feasible. Furthermore, the selection mechanism automatically filters unviable prompts from the population, given an adequate search budget. The keys to unlocking the full potential of evolutionary search are more targeted / fine-grained edit operations, structured search spaces, and surrogate-assisted fitness evaluations. In this context, the grammar-guided search \cite{whigham1995grammatically} offers a promising path forward.

\section{Approach}

\begin{table*}[t]
  \centering
    \caption{Edit operations that can be assigned by our grammar, divided into \textit{List Manipulation} and \textit{Semantically Aligned} operations. * \texttt{A/S} indicates an index can either be  $\bnfpn{\texttt{atomic_view}}$ or  $\bnfpn{\texttt{slice_view}}$; \texttt{A} indicates the index must be  $\bnfpn{\texttt{atomic_view}}$.}
  \label{tab:operations}
  \renewcommand{\arraystretch}{1.2} 
  \setlength{\tabcolsep}{6pt} 
  \begin{tabular}{p{0.08\textwidth} p{0.05\textwidth} p{0.25\textwidth} p{0.4\textwidth} p{0.1\textwidth}}
    \hline
    \textbf{Operator} & \textbf{Indices*}  & \textbf{Description} & \textbf{Example} & \textbf{Also used in} \\ \hline
    \multicolumn{5}{c}{\textsc{\textbf{List Manipulation Operations}} \textbf{$\bnfpn{\texttt{list_op}}$}} \\ \hline
    \texttt{swap} & 2$\times$\texttt{A/S} & Swaps the positions of element(s) at the indexed positions. & 
    [\underline{chunk\_1, chunk\_2}, chunk\_3, \underline{chunk\_4}] \newline $\rightarrow$  [chunk\_4, chunk\_3, chunk\_1, chunk\_2] & \cite{pan2024apoplum,prasad2022grips,zhang2022tempera} \\ \hline
    \texttt{remove} & 1$\times$\texttt{A/S} & Removes element(s) at index and stores in a FIFO queue. & 
    [chunk\_1, \underline{chunk\_2}, chunk\_3, chunk\_4] \newline $\rightarrow$  
    [chunk\_1, chunk\_3, chunk\_4] & \cite{pan2024apoplum,prasad2022grips,schnabel2024sammo,zhang2022tempera} \\ \hline
    \texttt{readd} & 1$\times$\texttt{A} & Adds previously removed element at index. & 
    [chunk\_1, \underline{chunk\_3}, chunk\_4] \newline $\rightarrow$  
    [chunk\_1, chunk\_2, chunk\_3, chunk\_4] & \cite{pan2024apoplum,prasad2022grips} \\ \hline
    \texttt{duplicate} & 1$\times$\texttt{A/S},\newline 1$\times$\texttt{A} & Inserts a duplicate copy of the chunk located at the source index to the target index. & 
    [\underline{chunk\_1, chunk\_2}, chunk\_3, \underline{chunk\_4}] \newline $\rightarrow$  
    [chunk\_1, chunk\_2, chunk\_3, chunk\_1, chunk\_2, chunk\_4] & \cite{pan2024apoplum,schnabel2024sammo,zhang2022tempera} \\ \hline
    \multicolumn{5}{c}{\textsc{\textbf{Semantically Aligned Operations}} \textbf{$\bnfpn{\texttt{semantic_op}}$}} \\  \hline
    \texttt{paraphrase} & 1$\times$\texttt{A/S} & Prompt-based invocation of an LLM to paraphrase NL text while maintaining semantics. & 
    ["Given text, classify its sentiment as positive or negative."] \newline $\rightarrow$  
    ["Is the sentiment of the given text positive or negative."] & \cite{cui2024phaseevo,pan2024apoplum,prasad2022grips,schnabel2024sammo} \\ \hline
    \texttt{summarise} & 1$\times$\texttt{A/S}  & Prompt-based invocation of an LLM to paraphrase text while limiting output to a certain percentage of the original length. & 
    ["Given text, classify its sentiment as positive or negative."] \newline $\rightarrow$  
    ["Classify the given text's sentiment."] & \cite{schnabel2024sammo} \\ \hline
    \texttt{rstopwords} & 1$\times$\texttt{A/S}  & Removes stop-words, as defined in the NLTK StopWords corpus\footnotemark[3]. & 
    ["Given text, classify its sentiment as positive or negative."] \newline $\rightarrow$  
    ["Given text, classify sentiment positive negative."] & \cite{schnabel2024sammo} \\ \hline
    \texttt{synonimise} & 1$\times$\texttt{A/S}  & Replaces words with synonyms retrieved from the WordNet corpus\footnotemark[4]. & 
    ["Given text, classify its sentiment as positive or negative."] \newline $\rightarrow$  
    ["Given text, categorise its sentiment as positive or negative."] & \cite{zhang2022tempera} \\ \hline
  \end{tabular}
\end{table*}

\subsection{Prompt Structure}
\label{method:structure}

To enable localised search of edit operations during optimisation, we partition the prompts into modular sections, allowing the search process to adaptively edit each functional part of the prompt without disrupting the entire template. We formalise the structure of a prompt template, $P$, as a composition of the following six sections, $s \in S$:

\begin{itemize}
    \item  \texttt{Persona}: Instructions for the LLM to adopt a specific persona or role while generating responses. 

    \item \texttt{Task}: Instructions specifying the particular task for the LLM to perform.
    \item \texttt{Output Format}: Instruction defining the required format of the output, often supplemented by output examples adhering to the specified structure.
    \item \texttt{In-Context Demonstrations}: For each case, five example demonstrations with the most similar inputs are retrieved from the training set to enable learning by a few shots \cite{brown2020fewshotlearners}. The grammar allows the search to determine which of these five are included and their order. 
    \item \texttt{Context}: A section reserved for additional retrieved information, typically when using retrieval-augmented generation (RAG). This includes a prefix describing the text followed by the information.
    \item \texttt{Chain-of-Thought (CoT)}: Instructions designed to invoke step-by-step reasoning in the model output, as proposed in \cite{kojima2023zeroshotcot}.
\end{itemize}

 Our grammar defines each of the \textit{Persona, Output Format, Context,} and \textit{CoT} sections as optional. This flexibility incorporates the ablation of these sections into our search space, enabling exploration of prompt variants that omit or modify specific structural components.

\subsection{Edit Operations}\label{sec:operators}


The grammar allows assignment of a range of edit operations, $\bnfpn{edit\_op}$, as listed in Tab. \ref{tab:operations}, which falls into two categories: 
\begin{enumerate}
    \item \textit{List Manipulation}, $\bnfpn{\texttt{list\_op}}$: Operations that manipulate the text as a list of chunked strings such as insertion, deletion, and reordering.
    \item \textit{Semantically Aligned Operations}, $\bnfpn{\texttt{semantic\_op}}$: Includes both LLM-based and text-processing operations designed to preserve or enhance semantic coherence. LLM-based edits, such as paraphrasing and summarising, are performed using the LLM. These edits are not applied to the In-Context Demonstrations.
\end{enumerate}

Each edit operation is invoked with two arguments: $\bnfpn{level}$, indicating the chunking granularity, and $\bnfpn{\texttt{index}}$, specifying the chunk(s) to edit. We define chunking as the non-desctructive segmentation of a string at a given \texttt{level} that allows reconstruction without any alterations. E.g. given the string `You are an expert. Answer this question:', chunking at the word \texttt{level} gives the list of chunks $C$: \texttt{['You', 'are', 'an', 'expert.', 'Answer', 'this', 'question:']} whereas chunking at the sentence \texttt{level} gives a $C$ of \texttt{['You are an expert.', 'Answer this question:']}.  After each edit,  all chunks are combined into a cohesive paragraph. In our approach, the grammar may assign the following chunking levels:
\begin{itemize}
    \item \texttt{Word}: tokenised with \texttt{NLTK.word\_tokenize()}.
    \item \texttt{Sentence}: split with \texttt{NLTK.sentence_tokenize()}.
    \item \texttt{Phrase}: parsed with a constituency-based CRF parser \footnote{\url{https://github.com/yzhangcs/parser}}.
\end{itemize} 

The $\bnfpn{\texttt{index}}$ may be an $\bnfpn{\texttt{atomic_view}}$ (a single integer) or a $\bnfpn{\texttt{slice_view}}$ index (two integers that define the start and end of a contiguous span). If a selected index $i$ exceeds the number of chunks in the input text, the edit is applied to the $(i \bmod |C|)$-th chunk. E.g. an atomic of \texttt{(5)} for the lists above returns the elements \texttt{'this'} and \texttt{'Answer the question:'}, respectively. 


\subsection{Grammar-Guided Genetic Programming (G3P)}\label{sec:optim}

\begin{algorithm}[t]
\caption{Grammar Model excerpt: the \textit{Persona} section.}\label{alg:grammar}

\begin{bnf*}
    \bnfprod{persona\_sec}{\bnfpn{persona\_expr} \bnfor \bnfpn{null\_string} }\\
    \bnfprod{persona\_expr}{\bnfpn{edit\_op} \bnfpn{persona\_expr} \bnfor \bnfpn{\bnfts{PERSONA}}} \\
    \bnfprod{edit\_op}{\bnfpn{list\_op} \bnfor \bnfpn{semantic\_op}} \\
    \bnfprod{PERSONA}{\bnfts{``You are an expert in..."}}\\
    \bnfprod{null\_string}{\bnfts{" "}}
\end{bnf*}
\end{algorithm}


\paragraph{Grammar:}\label{sec:grammar} Grammar-Guided Genetic Programming (G3P)~\cite{whigham1995grammatically} is a form of Genetic Programming (GP)~\cite{DBLP:books/daglib/0070933} that uses a formal grammar, typically in Backus–Naur Form, to guide the generation and variation operators (crossover and mutation) of expression-tree structures. The grammar constrains the search space and ensures syntactic correctness during evolution~\cite{DBLP:books/sp/18/NicolauA18}.

In our G3P DPO approach, our grammar $G$ consists of production rules that yield Python-executable code for applying $\bnfpn{\texttt{edit\_op}}$ transformations to sections, $P_s$, of a human-authored base prompt, $P_0$. See Appendix~\ref{app:grammar} for our complete grammar. In a given population, $\mathcal{P}$, each individual $I$ is represented by its genotype (a sequence of integers) that maps through $G$ to its phenotype -- a set of edit sequences $E_{S,I} = \{\,e_{s,I}\mid s \in S\}$,  one for each prompt section where $S = \{\texttt{Persona}, \texttt{Task}, \texttt{Output}, \texttt{ICL}, \texttt{Context}, \texttt{CoT}\}$ per Sec. \ref{method:structure}.  Each $e_{s,I}$ is executed on its corresponding section of the base prompt $P_s$. Each $e_{s,I}$ is applied independently to its corresponding $P_s$, enabling modular edits while preserving structural integrity, and the final prompt is given as:
\begin{equation}
P_{I} = \texttt{concatenate} \left( \texttt{execute}(e_{s,I}(P_s)) \right), \quad \forall s \in S.
\end{equation}

where \texttt{execute()} represents Python's built-in \texttt{eval} function and \texttt{concatenate()} joining multiple strings into one.

We provide a simplified example of our grammar in Example \ref{alg:grammar}. In this example, $\bnfpn{PERSONA}$ is the text of the \textit{Persona} section from $P_0$. The edited section $\bnfpn{persona_sec}$ may be defined as a null string or as $\bnfpn{persona\_expr}$, which recursively applies the edit operators $\bnfpn{\texttt{edit\_op}}$ to the original text from within the original prompt. We present the edit operators available in our grammar, $\bnfpn{\texttt{edit\_op}}$, in Sec. \ref{sec:operators}. See our complete grammar in Appendix \ref{app:grammar}.

\footnotetext[3]{Corpus 92 at \url{https://www.nltk.org/nltk_data/}}
\footnotetext[4]{\url{https://wordnet.princeton.edu/}}
\addtocounter{footnote}{2}

\paragraph{Grammar-Guided Genetic Programming:} To identify the optimum set of edit sequences $E_{S,I}$, we employ the Context-Free Grammar-Guided Genetic Programming on Serialised Trees algorithm as implemented in the Alogos framework\footnote{\url{https://github.com/robert-haas/alogos}}. Evolution proceeds through tournament selection, subtree crossover, and subtree mutation operators over a population $\mathcal{P}$ initialised with PTC2~\cite{luke2000two}. For $G$ generations, every individual in $\mathcal{P}$ is evaluated: their phenotype, $E_{S,I}$, is executed to create their prompt template, $P_I$. The fitness of each individual, $f_I$, is measured by prompting $M_T$ with $P_I$ on a random sample of the training data, $d_{\text{train}}\subset\mathcal{D}_{\text{train}}$.

The individual with maximum $f_I$ in each generation is validated on the validation set $\mathcal{D}_{\text{val}}$, giving that individual's validation fitness $f_{I,\text{val}}$. Each individual's fitness determines its likelihood of survival into the next generation and selection for reproduction via subtree crossover and mutation. Survivors and selected offspring form the updated population for the next generation. The settings and operators used for our search are presented in Appendix \ref{app:settings}.

The individual with maximum $f_{I,\text{val}}$ is deemed the elite $\mathcal{E}$. At the start of every generation, where $\mathcal{E}$ is not in $\mathcal{P}$, it replaces the worst performing individual $I_{\text{worst}} = \arg\min_{I \in \mathcal{P}} f_I$. After $G$ generations, the prevailing elite $\mathcal{E}$ produces the best individual $I^*$ with its prompt representing the final optimised prompt $P^*$. This prompt is then evaluated on a held-out testing set $\mathcal{D}_{\text{test}}$.

\subsection{Post-Hoc Local Search}
\label{sec:phlc}

\textbf{Neighbourhood:}
Starting from the best-evolved individual in G3P DPO, $I^*$ and its set of edit sequences, $E_{S,I}$, we define its neighbours using a point mutation operator which randomly perturbs an \texttt{index} parameter in $E_{S,I}$. For instance, applying this operator to \texttt{swap($idx_1$, $idx_2$)} gives either \texttt{swap($idx_3$, $idx_2$)} or \texttt{swap($idx_1$, $idx_3$)}, where $idx_1$ and $idx_2$ are the index values in $e_S$ and $idx_3 \in \{1,\ldots, U\}$. The upper bound $U$ is defined as $2\times$ the maximum number of chunks found in an input to an edit operation during the execution of $E_{S,I}$.

\begin{figure}[b]

    \centering
        \begin{minipage}[t]{\linewidth}
        \centering
        \resizebox{\linewidth}{!}{


\newcommand{\customshape}[2]{
    \draw[fill=white] 
        (#1-7.07, #2+15) -- (#1-7.04, #2+10.38)
        .. controls (#1-7.02, #2+8.88) and (#1-3.86, #2+6.87) .. (#1, #2+6.9)
        .. controls (#1+3.91, #2+6.93) and (#1+7.05, #2+8.97) .. (#1+7.03, #2+11.47)
        -- (#1+7, #2+15)
        -- cycle;
                
    \draw[fill=white] 
        (#1-3.5, #2+5) .. controls (#1-3.5, #2+2.51) and (#1-1.49, #2+0.5) .. (#1, #2+0.5)
        .. controls (#1+1.49, #2+0.5) and (#1+3.5, #2+2.51) .. (#1+3.5, #2+5)
        .. controls (#1+3.5, #2+7.49) and (#1+1.49, #2+9.5) .. (#1, #2+9.5)
        .. controls (#1-1.49, #2+9.5) and (#1-3.5, #2+7.49) .. (#1-3.5, #2+5)
        -- cycle;
}

\tikzset{every picture/.style={line width=0.75pt}} 

\begin{tikzpicture}[x=0.75pt,y=0.75pt,yscale=-1,xscale=1]

\draw[fill=black!20]   (100,100) -- (500,100) -- (500,370) -- (100,370) -- cycle ;

\draw[fill=white]    (2.7175 + 72.2175, 311) -- (2.7175 + 72.2175, 339) .. controls (2.7175 + 72.2175, 342.31) and (-13.1525 + 72.2175, 345) .. (-32.7125 + 72.2175, 345) .. controls (-52.1875 + 72.2175, 345) and (-67.1525 + 72.2175, 342.31) .. (-67.1525 + 72.2175, 339) -- (-67.1525 + 72.2175, 311) .. controls (-67.1525 + 72.2175, 307.68) and (-52.1875 + 72.2175, 304) .. (-32.7125 + 72.2175, 304) .. controls (-13.1525 + 72.2175, 304) and (2.7175 + 72.2175, 307.68) .. (2.7175 + 72.2175, 311) .. controls (2.7175 + 72.2175, 314.31) and (-13.1525 + 72.2175, 317) .. (-32.7125 + 72.2175, 317) .. controls (-52.1875 + 72.2175, 317) and (-67.1525 + 72.2175, 314.31) .. (-67.1525 + 72.2175, 311);

\draw[dashed,->] (74.935, 325)   -| (220, 270)  node[pos=0.75, left, align=left] {Train}  -|  (290, 260);

\draw[rounded corners=5pt, fill=black!80]  
(40 - 50, 120) -- (40 + 50, 120) -- (40 + 50, 170) -- (40 - 50, 170) -- cycle;

\draw[->] (90, 145) -- node[pos=0.5, above, align=left] {Phenotype} (240, 145);

\draw[fill=white]    (240,120) -- (340,120) -- (340,170) -- (240,170) -- cycle ;
\draw[->] (340, 145) -- (380, 145);
\draw[fill=white]  (380,120) -- (480,120) -- (480,170) -- (380,170) -- cycle;
\draw[->] (430, 170) |- (430, 190) -| (150, 210);

\draw[fill=white]  (0, 190 + 20) -- (80, 190 + 20) -- (80, 240 + 20) -- (0, 240 + 20) -- cycle;
\draw (40, 235) node [anchor=center] [inner sep=0.75pt] [align=center] {G3P\\DPO};
\draw[->] (40, 210) -- (40, 170);
\draw[->] (40, 260) -- (40, 304);

\customshape{40}{150}

\draw[rounded corners=5pt,fill=black!80]  (120, 190 + 20) -- (220, 190 + 20) -- (220, 240 + 20) -- (120, 240 + 20) -- cycle;
\draw[->] (220, 215 + 20) -- (240, 215 + 20);

\draw[fill=white]  (240, 190 + 20) -- (340, 190 + 20) -- (340, 240 + 20) -- (240, 240 + 20) -- cycle;
\draw[->] (340, 215 + 20) -- (380, 215 + 20);

\draw[fill=white]  (380, 190 + 20) -- (480, 190 + 20) -- (480, 240 + 20) -- (380, 240 + 20) -- cycle;
\draw[->] (430, 260) |- (430, 290) -| (290, 300);

\draw[fill=white]  (240, 260 + 40) -- (340, 260 + 40) -- (340, 310 + 40) -- (240, 310 + 40) -- cycle;
\draw[->] (340, 325) -- node[above,align=left] {Top-1} (380, 325);

\draw[rounded corners=5pt,fill=black!80]  (380, 260 + 40) -- (480, 260 + 40) -- (480, 310 + 40) -- (380, 310 + 40) -- cycle;

\draw (40, 135) node [anchor=center] [inner sep=0.75pt] [align=center] {\small\textcolor{white}{Best Individual}};

\draw (290, 145) node [anchor=center] [inner sep=0.75pt] [align=center] {Extract All\\Indices};

\draw (430, 145) node [anchor=center] [inner sep=0.75pt] [align=center] {Randomly\\Select Value\\ \footnotesize (10 per index)};

\draw (170, 225) node [anchor=center] [inner sep=0.75pt] [align=center] {\textcolor{white}{Neighbours}};
\customshape{135}{237}
\draw (140,240) node [anchor=north west][inner sep=0.75pt]   [align=left] {{ {\small \textcolor{white}{$\times10$ per index}}}};

\draw (290, 215 + 20) node [anchor=center] [inner sep=0.75pt] [align=center] {Surrogate\\Model};

\draw (430, 215 + 20) node [anchor=center] [inner sep=0.75pt] [align=center] {Neighbours\\ \footnotesize (Top-25 Mean \& \\\footnotesize Top-25 Std)};

\draw (290, 285 + 40) node [anchor=center] [inner sep=0.75pt] [align=center] {Eval on\\Validation Set};

\draw (430, 285 + 30) node [anchor=center] [inner sep=0.75pt] [align=center] {\textcolor{white}{Best Individual}};
\customshape{430}{285 + 40}
\draw (40, 343) node [anchor=south] [inner sep=0.75pt] [align=center] {\small Training\\Fitness};

\end{tikzpicture}

        }

        \caption{Our Post-Hoc Local Search, which generates and evaluates variations of the best individual found during G3P DPO, each differing from one another by the value of one index.}
        \label{fig:local}
        
    \end{minipage}
    %
\end{figure}
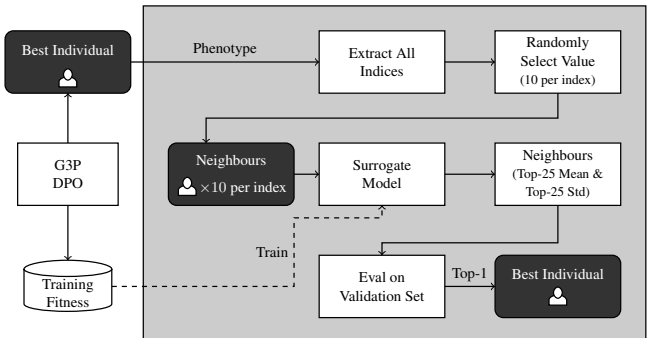

Using this operator, we enumerate the neighbourhood $N_{I^*}$ of $I^*$ by targetedly applying it to every index within $E_{S,I}$. We create 10 unique neighbours for each index, resulting in a neighbourhood size of $|N_{I^*}| = 10\times n_{idx,I}$, where $n_{idx,I}$ is the number of indexes in $E_{S,I}$. The fitness of each neighbour is predicted using our ensemble-based surrogate model which returns the mean and variance of each neighbour's predicted fitness across its 10 submodels. To balance exploration and exploitation, we select the 25 neighbours with the highest mean and the 25 with the highest variance of the predicted fitness, $\hat{f}(P)$ and $\sigma^2(P)$ respectively. The resulting 50 neighbours are then scored by combining their performance on the validation set $D_{\text{val}}$ and on a random sample $d_{\text{train}}\subset D_{\text{train}}$ with $|d_{\text{train}}|=|D_{\text{val}}|$. The neighbour with the highest combined score is retained as our best individual, yielding the optimized prompt. We illustrate this local search process in Fig.~\ref{fig:local}.

\textbf{Surrogate Model:}\label{method:surrogate}
To manage the computational costs of local search over large neighbourhoods, we train a surrogate ensemble to predict prompt fitness from a Sentence-BERT embedding, using data collected during G3P DPO.

We employ an ensemble of neural networks $F_i: X_P \in \mathbb{R}^{1\times D} \rightarrow f \in \mathbb{R}$, where $X_P$ is the embedding of prompt $P$, $D$ is the length of $X_P$, $f$ is the fitness of $P$, and $i \in \{1,\ldots,10\}$. Each $F_i$ comprises of linear layers of decreasing widths connected with tanh activation and dropout. To train the ensemble, we first split the training data in a 70:30 ratio into $D_{\mathrm{surr,train}}$ and $D_{\mathrm{surr,val}}$, respectively. We then use bootstrap sampling, with replacement, on $D_{\mathrm{surr,train}}$ to create training subsets $d_{\mathrm{surr,train}}^i$. Each model $F_i$ is trained on $d_{\mathrm{surr,train}}^i$ for $T = 200$ epochs. Let $\mathcal{L}_i^{(t)}$ denote the validation loss of model $F_i$ at epoch $t$. We track the average ensemble validation loss across all submodels as $\bar{\mathcal{L}}^{(t)} = \tfrac{1}{10}\sum_{i=1}^{10}\mathcal{L}_i^{(t)}$ and store the parameters of the models $W_{t}$ corresponding to the epoch $t^*$ such that $t^* = \arg\min_{t\in\{1,\ldots,T\}}\bar{\mathcal{L}}^{(t)}$. 

The learning rate, batch size, neural network topology, and dropout rate are defined for the sorrugated ensemble during initialisation by way of hyperparameter testing. We sample 10 unique combinations of these hyperparameters and test each using a 5-fold cross-validation of the mean square error between the predicted and actual fitness values as averaged across all submodels. We then used the best-performing hyperparameter set to train the surrogate model. 

At inference time, the ensemble with parameters $W_{t}$ takes in $X_P$ of each neighbour in $N_{I^{*}}$ and predicts its fitness $F_i(X_P)$ for each model $i \in {1,\ldots,10}$. The surrogate returns the mean predicted fitness $\hat{f}(P) = \tfrac{1}{10}\sum_{i=1}^{10}F_i(X_P)$, and the variance, $\sigma^2(P) = \tfrac{1}{10}\sum_{i=1}^{10}\bigl(F_i(X_P)-\hat{f}(P)\bigr)^2$. $\hat{f}(P)$ represents the ensemble's estimate of the neighbour’s fitness, while $\sigma^2(P)$ provides an estimate of the epistemic uncertainty associated with the surrogate’s prediction.


\section{Evaluation}
\label{sec:evaluation}

\begin{table}[t]
\caption{Datasets used in our experiments. Performance for each task is reported using the metric defined by each task's respective authors.}
\label{tab:tasks}
\centering
\begin{small}
\begin{tabular}{p{0.15\linewidth} p{0.25\linewidth} p{0.25\linewidth} p{0.2\linewidth}}
\toprule
\textbf{Dataset} & \textbf{Task\newline Description} & \textbf{Output} & \textbf{Metric} \\
\midrule
\textbf{PubMedQA} 
&\begin{tabular}{@{}l@{}}Biomedical\\question answering\\on research papers. \end{tabular}
& \begin{tabular}{@{}l@{}}\texttt{yes / no /}\\\texttt{maybe}\end{tabular} &\begin{tabular}{@{}l@{}}Classification\\Accuracy\end{tabular}\\
\midrule
\textbf{ETHOS} 
& \begin{tabular}{@{}l@{}}Binary hate speech\\classification\end{tabular} 
& \begin{tabular}{@{}l@{}}\texttt{hateful} /\\\texttt{not hateful}\end{tabular} & \begin{tabular}{@{}l@{}}Classification\\Accuracy\end{tabular} \\
\midrule
\textbf{TAT-QA }
& \begin{tabular}{@{}l@{}}Numerical question\\answering\\on tabular data\end{tabular}
& \begin{tabular}{@{}l@{}}Various incl.:\\open-ended text,\\arithmetic\\formula and dates.\end{tabular} & F1 \\
\midrule
\textbf{ConvFinQA} 
& \begin{tabular}{@{}l@{}}Financial question\\answering on\\conversational and\\tabular data.\end{tabular} 
& \begin{tabular}{@{}l@{}}Executable\\numerical\\formula.\end{tabular}  & \begin{tabular}{@{}l@{}}Formula\\Execution\\Accuracy\end{tabular} \\
\bottomrule
\end{tabular}
\end{small}
\end{table}

\subsection{Datasets, Models, and Prompts}

\textbf{Datasets:} Four challenging and domain-specific NLP tasks are selected to evaluate the proposed approach: \textbf{PubMedQA}, a biomedical question-answering dataset collected from PubMed abstracts requiring domain-specific reasoning \cite{jin2019pubmedqa}; \textbf{ETHOS}, a binary classification dataset requiring detection of hate speech and offensive language in text \cite{mollas2022ethos}; \textbf{ConvFinQA}, a conversational financial question-answering dataset designed for multiturn numerical reasoning \cite{chen2022convfinqa}; and \textbf{TAT-QA}, a large-scale question-answering dataset requiring reasoning over tabular and textual data \cite{zhu2021tat}. These datasets provide a broad testbed covering different types of reasoning, multiple domains, and diverse language understanding tasks. 

To ensure consistency with evaluation protocols adopted by the benchmarks, we reuse the train-validation-test data split for \textbf{TAT-QA} and the train-dev split for \textbf{ConvFinQA}. For the latter, the dev set is split 50:50 to form validation and test sets. We also reuse the author-defined splits for \textbf{PubMedQA} with the sampling and randomisation seed provided by the authors. Since \textbf{ETHOS} does not have predefined splits, random sampling is used to form splits with a 50:20:30 train-val-test ratio. The same splits are maintained when evaluating all approaches, models, and methods. See Appendix \ref{app:tasks} for additional details on the datasets. The performance of a prompt is measured using the corresponding evaluation metrics as defined by the authors of each dataset. The performance of a prompt on a given task is measured using syntactic comparison to ground-truth labels provided in the dataset. This eliminates the need to use LLMs to score its own output.

\begin{table*}[!t]
\caption{Performance of prompts optimised using benchmark and our approaches. Relative gain over baseline prompt performance shown in parantheses. Best and second-best performance measured for each task-model combinantion is bolded and underlined, respectively. *Qwen2.5 32B was only used to evaluate the performance of OPRO with a larger model. **Excludes Qwen2.5 32B results for comparability with other approaches.}
\normalsize
\label{tab:results}
\centering
\begin{tabular}{>{\arraybackslash}m{0.09\textwidth} >{\raggedleft\arraybackslash}m{0.05\textwidth} >{\centering\arraybackslash}m{0.1\textwidth} >{\centering\arraybackslash}m{0.1\textwidth} >{\centering\arraybackslash}m{0.1\textwidth} >{\centering\arraybackslash}m{0.1\textwidth} >{\centering\arraybackslash}m{0.1\textwidth} >{\centering\arraybackslash}m{0.1\textwidth} }
\toprule

 &  & &  \multicolumn{3}{c}{\textbf{Benchmark Approaches}} & \multicolumn{2}{c}{\textbf{Our Approaches}} \\
 \textbf{Model} & \textbf{Size} & \textbf{Baseline Prompt} & \textbf{PromptWizard} & \textbf{OPRO} & \textbf{RL-Prompt} & \textbf{G3P DPO} & \textbf{G3P DPO + Local Search} \\
\hline
\multicolumn{8}{c}{\textbf{PubMedQA}} \\
\hline
\textbf{Llama3.2}&\textbf{3B} & 43.8 & 56.8 \footnotesize{ (\textuparrow30\%)} & 29.5 \footnotesize{\hphantom{0}(\textdownarrow33\%)} & 10.5 \footnotesize{\hphantom{0}(\textdownarrow76\%)} & \textbf{70.0} \footnotesize{\hphantom{0}(\textuparrow\textbf{60\%})} & \underline{66.5} \footnotesize{\hphantom{0}(\textuparrow 52\%)} \\
\textbf{Llama3}&\textbf{8B}  & 31.7 & \textbf{72.2} \footnotesize{(\textuparrow \textbf{128\%})} & 34.5 \footnotesize{\hphantom{00}(\textuparrow 9\%)} & 40.0 \footnotesize{\hphantom{0}(\textuparrow 26\%)} & 67.3 \footnotesize{(\textuparrow 112\%)} & \underline{69.7} \footnotesize{(\textuparrow 120\%)} \\
\textbf{Gemma2}&\textbf{9B}  & 58.2 & 34.2 \footnotesize{\hphantom{0}(\textdownarrow41\%)} & 28.41 \footnotesize{\hphantom{0}(\textdownarrow51\%)} & 26.5 \footnotesize{\hphantom{0}(\textdownarrow55\%)} & \underline{61.8} \footnotesize{\hphantom{00}(\textuparrow 6\%)} & \textbf{64.8} \footnotesize{\hphantom{0}(\textuparrow \textbf{11\%})} \\
\textbf{Qwen2.5}$^{*}$&\textbf{32B}  & \textbf{70.4} & -- & 56.06 \footnotesize{\hphantom{0}(\textdownarrow20\%)} & -- & -- & -- \\
\hline
\multicolumn{8}{c}{\textbf{ETHOS}} \\
\hline
\textbf{Llama3.2}&\textbf{3B} & 70.3 & \textbf{78.0} \footnotesize{\hphantom{0}(\textuparrow \textbf{11\%})} & 59.7 \footnotesize{\hphantom{0}(\textdownarrow15\%)} & 25.0 \footnotesize{\hphantom{0}(\textdownarrow64\%)} & 71.0 \footnotesize{\hphantom{00}(\textuparrow 1\%)} & \underline{77.3} \footnotesize{\hphantom{0}(\textuparrow 10\%)} \\
\textbf{Llama3}&\textbf{8B}  & 39.1 & 58.7 \footnotesize{\hphantom{0}(\textuparrow 50\%)} & 27.6 \footnotesize{\hphantom{0}(\textdownarrow29\%)} & \hphantom{0}4.7 \footnotesize{\hphantom{0}(\textdownarrow88\%)} & \underline{73.3} \footnotesize{\hphantom{0}(\textuparrow 88\%)} & \textbf{81.3} \footnotesize{(\textuparrow \textbf{108\%})} \\
\textbf{Gemma2}&\textbf{9B}  & 83.0 & 68.0\footnotesize{\hphantom{0}(\textdownarrow18\%)} & \hphantom{0}0.0 \footnotesize{\hphantom{0}(\textdownarrow100\%)} & \underline{83.5} \footnotesize{\hphantom{00}(\textuparrow 1\%)} & 83.3 \footnotesize{\hphantom{00}(\textuparrow 0\%)} & \textbf{83.7} \footnotesize{\hphantom{00}(\textuparrow \textbf{1\%})} \\
\textbf{Qwen2.5}$^{*}$&\textbf{32B}  & 89.0 & -- & \textbf{89.7} \footnotesize{\hphantom{00}(\textuparrow 1\%)} & -- & -- & -- \\
\hline
\multicolumn{8}{c}{\textbf{TAT-QA}} \\
\hline
\textbf{Llama3.2}&\textbf{3B} & 24.9 &\hphantom{0}2.9 \footnotesize{\hphantom{0}(\textdownarrow88\%)} & 35.0 \footnotesize{\hphantom{0}(\textuparrow 41\%)} & 21.8 \footnotesize{\hphantom{0}(\textdownarrow12\%)} & \underline{42.7} \footnotesize{\hphantom{0}(\textuparrow 72\%)} & \textbf{44.2} \footnotesize{\hphantom{0}(\textuparrow \textbf{78\%})} \\
\textbf{Llama3}&\textbf{8B}  & 10.9 & \textbf{36.4} \footnotesize{(\textuparrow \textbf{234\%})} & 29.5 \footnotesize{(\textuparrow 171\%)} & \hphantom{0}5.5 \footnotesize{\hphantom{0}(\textdownarrow49\%)} & 32.3 \footnotesize{(\textuparrow 197\%)} & \underline{34.6} \footnotesize{(\textuparrow 218\%)} \\
\textbf{Gemma2}&\textbf{9B}  & \underline{40.5} & 38.7 \footnotesize{\hphantom{00}(\textdownarrow4\%)} & 38.9 \footnotesize{\hphantom{0}(\textdownarrow4\%)} & 34.5 \footnotesize{\hphantom{0}(\textdownarrow15\%)} & 34.8 \footnotesize{\hphantom{0}(\textdownarrow14\%)} & \textbf{40.6} \footnotesize{\hphantom{00}(\textuparrow \textbf{0\%})} \\
\textbf{Qwen2.5}$^{*}$&\textbf{32B}  & 37.0 & -- & \textbf{63.2} \footnotesize{\hphantom{0}(\textuparrow 71\%)} & -- & -- & -- \\
\hline

\multicolumn{8}{c}{\textbf{ConvFin-QA}} \\
\hline
\textbf{Llama3.2}&\textbf{3B} & 34.5 & 18.2 \footnotesize{\hphantom{0}(\textdownarrow47\%)} & 17.8 \footnotesize{\hphantom{0}(\textdownarrow48\%)} & \hphantom{0}3.6 \footnotesize{\hphantom{0}(\textdownarrow90\%)} & \underline{38.4} \footnotesize{\hphantom{0}(\textuparrow 11\%)} & \textbf{38.8} \footnotesize{\hphantom{0}(\textuparrow \textbf{12\%})} \\
\textbf{Llama3}&\textbf{8B}  & 19.6 & \hphantom{0}1.4 \footnotesize{\hphantom{0}(\textdownarrow93\%)} & \hphantom{0}5.5 \footnotesize{\hphantom{0}(\textdownarrow72\%)} & \hphantom{0}8.5 \footnotesize{\hphantom{0}(\textdownarrow57\%)} & \underline{23.1} \footnotesize{\hphantom{0}(\textuparrow 18\%)} & \textbf{33.5} \footnotesize{(\hphantom{0}\textuparrow \textbf{71\%})} \\
\textbf{Gemma2}&\textbf{9B}  & \textbf{61.9} & \hphantom{0}1.2 \footnotesize{\hphantom{0}(\textdownarrow98\%)} & 53.6 \footnotesize{\hphantom{0}(\textdownarrow13\%)} & 45.5 \footnotesize{\hphantom{0}(\textdownarrow26\%)} & 49.8 \footnotesize{\hphantom{0}(\textdownarrow20\%)} & \underline{54.5} \footnotesize{\hphantom{0}(\textdownarrow12\%)} \\
\textbf{Qwen2.5}$^{*}$&\textbf{32B}  & 37.0 & -- & \textbf{69.4} \footnotesize{\hphantom{0}(\textuparrow 88\%)} & -- & -- & -- \\
\hline
\multicolumn{2}{l}{\textbf{Mean Relative Gain}} & -- &\textuparrow5\% &\textdownarrow12\%$^{**}$ & \textdownarrow42\% & \textuparrow44\%&\textuparrow\textbf{56\%} \\
\hline
\end{tabular}
\end{table*}

\textbf{Models:} We evaluate three instruction-tuned LLMs of varying sizes for both prompt editing and generation tasks: \textbf{Llama3.2 3B}, \textbf{Llama3 8B}, and \textbf{Gemma2 9B}; see Appendix \ref{app:llmsettings} for LLM hyperparameters. These intermediate size LLMs strike a balance between general-purpose reasoning ability and computational efficiency. Clear and concise task instructions are more crucial for smaller LLMs due to their lack of domain-specific knowledge, attenuated capabilities in reasoning tasks \cite{li2025smallmodelsstrugglelearn, emergentlarge}, and to deduce task requirements from provided demonstrations \cite{min-etal-2022-rethinking}.

\textbf{Prompts:} For each task, we establish baseline performance by taking a manually engineered baseline prompt template and providing five in-context demonstrations as described in Sec \ref{method:structure}; see baseline prompt templates for each task in Appendix \ref{app:base-templates}. Optimisation of these prompts proceeds in two steps. First, the baseline prompt is optimised using the G3P DPO algorithm described in Section \ref{sec:grammar}. Second, the partially optimised prompt from G3P DPO seeds the Post-Hoc Local Search algorithm described in Section \ref{sec:phlc}. To assess the benchmark algorithms, the baseline prompt is provided as input to PromptWizard, OPRO and RL-Prompt. Finally, the performance of the baseline prompts and optimised prompts realised by all methods are assessed on the holdout test set of all four datasets.


The tasks we evaluate are challenging, especially for smaller LLMs. Prior work applied a wide range of models. Consequently, comparable baseline performance results for the models considered in this paper are not available. Nonetheless, to provide some context, we report both the literature baseline performance results in Appendix \ref{app:baselines}, and the performance of our models using baseline prompts. Overall, our results are in line with existing performance trends by model size.

\subsection{Benchmark Approaches}

We benchmark our approach against three leading prompt optimisation methods: \textbf{PromptWizard} \cite{agarwal2024promptwizard}, \textbf{OPRO} \cite{yang2024opro}, and \textbf{RL-Prompt} \cite{deng2022rlprompt}. PromptWizard was run with the default settings defined by its authors. 

To manage the significantly higher costs of OPRO and RL-Prompt, we limit the LLM-based evaluations on training data to match that of our approach. We did so to both methods by setting the number of training data per step to 10 rows, randomly sampled for each step. For RL-Prompt we also reduced the number of candidate prompts per step to 10
Additionally, OPRO, by default, rejected prompts longer than 500 characters; we increased this limit to 5000. Finally, we updated the RL-Prompt baseline by appending the searched tokens to the baseline prompt -- rather than replacing the entire prompt. In our preliminary tests, the short prompts only yielded zero or near-zero performance across all tasks. All other OPRO and RL-Prompt settings follow the respective authors’ recommendations.


\section{Results}

\subsection{Overview}
Tab. \ref{tab:results} shows the performance of the baseline prompt, each benchmark, and our two approaches. We found that the G3P DPO optimised prompts outperformed the baseline prompts in 10 of 12 task-model combinations and Local Search added a further performance improvement in 11 of 12 combinations. We found G3P + Local Search to be the best or second-best for all combinations tested. Of the benchmark approaches, PromptWizard performed best; improving over baseline prompt performance in 6 cases and outperforming both our approaches in 3 cases. This shows that while PromptWizard can improve prompt performance with smaller models, its effectiveness was inconsistent across tasks and models.

Furthermore, when applying our approach to LLMs with different capabilities (as observed by the difference in performance figures on the baseline prompt), we observed a smaller \textit{inter-model spread} i.e., the difference between the performance of the best and worst models, than that of the baseline prompts. Specifically, the baseline prompts had an average inter-model spread of 35.6 across the four tasks. This was reduced to 14.4 with G3P DPO and 10.5 with G3P + Local Search. This reduction was mainly due to the performance gains for the weakest model. In particular, Llama3 8B was the weakest performer on all tasks with the baseline prompts, and experienced the largest relative improvement with our approaches. In contrast, the benchmark approaches showed larger performance degradations which led to greater average spreads of 27.5, 30.8, and 44.8 for PromptWizard, OPRO, and RL-Prompt, respectively. A smaller spread between models could reduce the penalties associated with selecting lower-performance models, allowing the use of these smaller and more efficient models with less compromise in performance.

On average, the prompts optimised with G3P DPO used 35.4 edit operations; comprising 32\%, 28\%, and 40\% LLM-based and non-LLM-based $\bnfpn{\texttt{semantic\_op}}$, and $\bnfpn{\texttt{list\_op}}$ operations, respectively.
Of these, 36\% were performed at the \texttt{word} level, and 43\% and 21\% at the \texttt{phrase} and \texttt{sentence} levels, respectively. Among the 12 optimised prompts, 7 had at least one section that was not edited. Furthermore, while the grammar allows certain sections of the prompt to be removed, this was not used in any of the prompts. Taken together, these suggest that where the original text of a baseline prompt section is already optimal, our approach does not apply any unnecessary or detrimental edits.

\subsection{Low-Performance Prompts}
\begin{figure}[b]
    \centering
    \includegraphics[width=\linewidth]{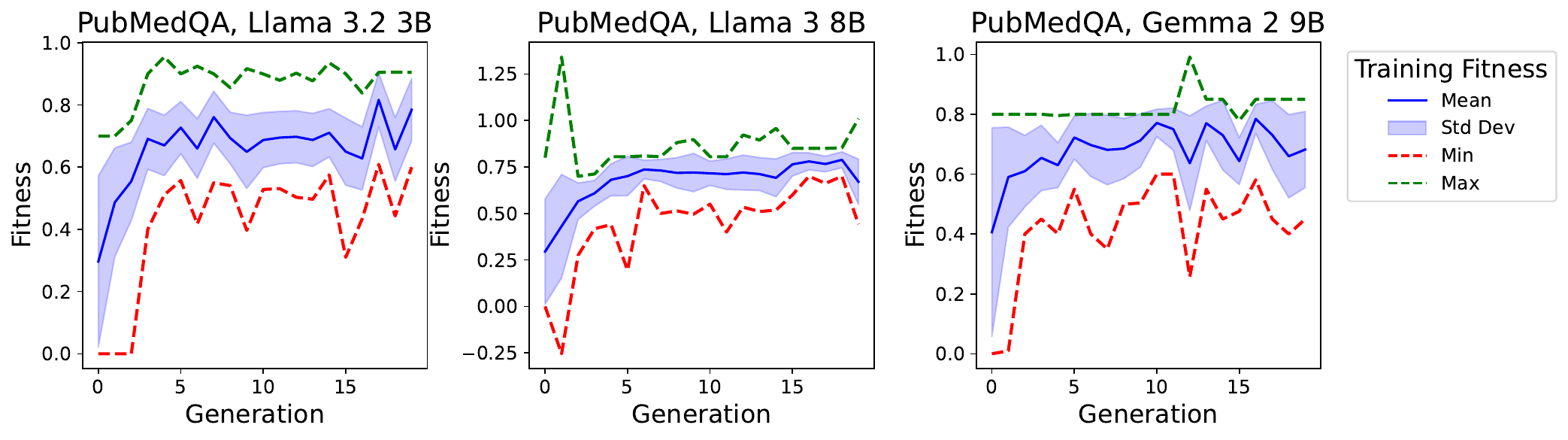}
    \caption{Training fitnesses measured during G3P DPO for the PubMedQA task. See similar plots for all tasks in Appendix \ref{app:learningcurves}.}
    \label{fig:trainingfit}
\end{figure}
Across the three benchmark approaches, we investigated several instances in which the performance of the optimised prompt was found to be severely degraded. In line with PromptWizard's authors caution against the use of models with fewer than 70 billion parameters with their technique, our attempt to do so revealed several failure modes. 

\textbf{PromptWizard:} The PromptWizard prompt for TAT-QA in Llama3.2 3B included an instruction that read ``\texttt{What is the change in net sales for cheese between 2018 and 2019?}''. This change appears to have stemmed from an attempt to induce prompt instruction using a particular sample of the training data. This inclusion led to the LLM attempting to answer this question regardless of the question for the given case at hand. For the ConvFinQA task, we observed PromptWizard's prompt template to include a heavily edited ICL demonstration which did not comply with the stated output format requirements; causing the LLM outputs during evaluation to almost always be unparseable. Although PromptWizard was shown to improve the performance of the ETHOS task in its original publication \cite{agarwal2024promptwizard}, we found that when applied with a small model to optimise and a longer prompt, it did not show the same efficacy.

\textbf{OPRO:} We investigated OPRO's prompt for the ETHOS task on Gemma 2 9B -- even rerunning this three more times gave the same performance figure. In this task, the outputs are restricted to two predefined classification labels but in OPRO's prompt these labels were converted to unrelated invalid values: ``\texttt{Positive}'' and ``\texttt{Negative}''. This caused the LLM to output invalid predictions during the evaluation. Suspecting that this and other performance issues with OPRO stemmed from the sizes of our selected model, we apply OPRO to optimise the prompt for all tasks in Qwen2.5 32B; the results are indicated with a * in Tab. \ref{tab:results}. With this larger and more capable model, OPRO prompts showed a stark increase in performance over respective baseline prompts in TAT-QA and ConvFinQA, a small improvement for ETHOS, and a less severe degradation for PubMedQA than was observed with smaller models. 

The sizable variance in OPRO's efficacy across different models requires that its application be preceded by testing to ensure that the chosen model is one that enables OPRO to improve a given prompt. As the capability and sizes of available LLMs change with the release of new open-source general-purpose models, the dependence of OPRO on certain (and likely unstated) capabilities of the LLM necessitates extensive testing to establish its compatibility with the model. This observation is in line with recent experiments showing that relatively small models cannot be self-optimised using OPRO \cite{min-etal-2022-rethinking}. 

\textbf{RL-Prompt:} Unlike the other two benchmark approaches, RL-Prompt does not rely on the LLM to edit the prompts. Instead, it searches for tokens to append to the baseline prompt. We suspect that the poor performance observed here may be due to two issues. First, in most of its authors' own evaluations, those tokens formed the entire prompt template; which is a setting suitable only for tasks needing less detailed instruction. During preliminary testing of this default approach, we found that the resulting prompt only led to zero or near-zero performance. However, since the number of LLM-based evaluations increases with the number of searched tokens compared to the length of the baseline prompt templates, increasing the search token count to that of the baseline prompts would have proved to be unfeasible. Secondly, in pursuing comparable compute costs, the reduced number of steps for which RL-Prompt was allowed to run was possibly too few to effectively apply the approach. In its original publications, the models tested were much smaller. Given the added computational costs from our relatively larger models, we were unable to use as many steps as the authors recommended or search for as many tokens as in the baseline prompts.

\textbf{G3P DPO:} The prompts generated by our solution are not immune to these failure modes, but the pressure of evolutionary selection and elitism make it unlikely that they appear in the final prompt. In early generations, we often observed candidates with zero training fitness; tournament selection disfavors their including genomes in subsequent populations. This can be observed via the distribution of training fitnesses of the individuals across the generations of our search; see training fitness curves for our G3P DPO in Fig. \ref{fig:trainingfit} and Appendix \ref{app:learningcurves}. Furthermore, only the highest scoring prompt from each generation is ever eligible for final selection. Effectively, prompts that cause the LLM to struggle with output formatting and classification labels (which typically show single-digit performance values) may only be selected as the final prompt where all generated prompts also contain similar or worse issues. Finally, our implementation of elitism guarantees that the genes of the best performing individual found at any point in the search, the elite, always remain in the population. This allows for further exploration of the search space around that individual while ensuring that if no better individual is found, then this elite will be returned as the final optimised prompt.

\subsection{Limitations \& Future Works}
While our method demonstrates strong empirical performance in a variety of combinations of tasks and models, it comes with a few inherent limitations and opens several avenues for future exploration. First, unlike some other DPO methods using LLM-based editing, our solution currently
does not include edit operations that introduce new information into the prompt,
such as data-driven instruction induction \cite{agarwal2024promptwizard}, self-reflection \cite{yang2024opro}, and phrase injection \cite{zhang2024sprig}. 

Furthermore, our evaluation focusses on selected structured NLP tasks and relatively small open-source language models. A promising future direction is to extend our framework to support prompt optimisation for visuo-lingual tasks, where inputs span both text and other modalities like images or videos. Additionally, incorporating embedding-based or semantic-aware edit operations may allow for more nuanced modifications beyond structural changes.
\section{Conclusion}
In this work, we addressed the problem of optimising prompts for smaller general-purpose LLMs on domain-specific tasks with detailed templates, demonstrating that existing DPO methods frequently fail in this setting. In response, we proposed a grammar-guided evolutionary search DPO solution a grammar model that defines sequences of edit operations on sections of a manually crafted prompt. We show that this approach improved the performance of three relatively small LLMs of varying sizes on four datasets; with only two model-task combinations where it was not able to. We also introduced a local search that further improves over the optimised prompts in all cases but one. We believe this approach will enable practitioners to tailor compact LLMs to specialised domains more reliably, and motivate future works to tackle DPO in this challenging scenario.

\bibliography{main}
\newpage
\onecolumn
\section*{Appendices}
\appendix

\section{Dataset Details}\label{app:tasks}
\begin{table}[h!]
\caption{Task Dataset Details}
\label{tab:datasetdetails}
\centering
\begin{tabular}{p{0.25\linewidth} p{0.13\linewidth} p{0.13\linewidth} p{0.13\linewidth} p{0.13\linewidth}}
\hline
\textbf{Dataset} & \textbf{PubMedQA}\cite{jin2019pubmedqa}  & \textbf{ETHOS}\cite{mollas2022ethos}  & \textbf{TAT-QA}\cite{tatllama}  & \textbf{ConvFinQA}\cite{chen2022convfinqa} \\
\hline
Training Set Size & 336 & 548 & 13,215 & 3,037 \\
Validation Set Size  & 164 & 150 & 1,668 & 210 \\
Testing Set Size  & 500 & 300 & 1,663 & 211 \\
Splits Provided by Dataset Authors  & Yes (via authors' splitting code) & No & Yes & Yes (train-dev split) \\
Task Selected (if more than one exists) & \texttt{PQA-Labeled} & Binary Classification & N/A & \texttt{Conversation}\\
Other available metrics & F1 & Precision, Recall, F1 & Exact Match & Program Accuracy\\
\hline
\end{tabular}
\end{table}

\section{Survey of Task Performance across Various Models}\label{app:baselines}

\begin{table}[!h]
\caption{Performance reported by previous works on each task using models of various sizes. * indicates that the model size is approximated due to it not being publicly disclosed.}
\label{tab:baselines}
\centering
\begin{tabular}{p{0.15\columnwidth} p{0.25\columnwidth} p{0.1\columnwidth} p{0.15\columnwidth} p{0.1\columnwidth}}
\toprule
\textbf{Dataset} & \textbf{Model} & \textbf{Size} & \begin{tabular}{@{}l@{}}\textbf{Reported}\\\textbf{Perf.}\\ \end{tabular} & \textbf{Source} \\
\midrule
\textbf{PubMedQA} 
&Flan-PaLM
& 540B & 79.0 & \cite{pubmedflan}\\
&GPT-3.5-turbo
& 175B* & 37.2 & \cite{pubmedgptturbo}\\
&Llama2 (finetuned)
& 7B & 59.4 & \cite{pubmedgptturbo}\\
&Llama3.2
& 3B & 43.8 & Our baseline\\
&Llama3
& 8B & 31.7 & Our baseline\\
&Gemma2 
& 9B & 58.2 & Our baseline\\
& 
& 32B & 70.4 & Our baseline\\

\midrule
\textbf{ETHOS} 
&\begin{tabular}{@{}l@{}}GPT-4 (w/ prompt\\optimisation)\end{tabular} 
& 1.8T* & 89.4 & \cite{agarwal2024promptwizard}\\
&Llama2
& 7B & 12.0 & \cite{ethos7b}\\
&GPT-J
& 6B & 48.4 & \cite{ethosgptj}\\
&Llama3.2
& 3B & 70.3 & Our baseline\\
&Llama3
& 8B & 39.1 & Our baseline\\
&Gemma2 
& 9B & 83.0 & Our baseline\\
& Qwen 
& 32B & 89.7 & Our baseline\\
\midrule
\textbf{TAT-QA }
&GPT-4
& 1.8T* & 75.27 & \cite{tatgpt4}\\
&Llama
& 70B & 45.5 & \cite{tatgpt4}\\
&Llama2
& 13B & 6.8 & \cite{tatllama}\\
&Llama2
& 7B & 22.4 & \cite{tatllama}\\
& Mixtral
& 7B & 7.6 & \cite{tatllama}\\
&Llama3.2
& 3B & 24.9 & Our baseline\\
&Llama3
& 8B & 10.9 & Our baseline\\
&Gemma2 
& 9B & 40.5 & Our baseline\\
& Qwen 
& 32B & 63.2 & Our baseline\\
\midrule
\textbf{ConvFinQA} 
&GPT-3
& 175B & 40.6 & \cite{chen2022convfinqa}\\
&Wizard-LM
& 13B & 29.9 & \cite{tatllama}\\
&Llama2
& 13B & 3.7 & \cite{tatllama}\\
&Llama2
& 7B & 7.5 & \cite{tatllama}\\
& Mixtral
& 7B & 7.6 & \cite{tatllama}\\
&Llama3.2
& 3B & 34.5 & Our baseline\\
&Llama3
& 8B & 19.6 & Our baseline\\
&Gemma2 
& 9B & 61.9 & Our baseline\\
& Qwen 
& 32B & 69.4 & Our baseline\\
\bottomrule
\end{tabular}
\end{table}

\section{Our Grammar}\label{app:grammar}
\begin{bnf*}
\bnfprod{PROMPT}{
\bnfpn{sec\_persona}, \bnfpn{sec\_task}, \bnfpn{sec\_task\_cons}, \bnfpn{sec\_icl}, }\\
\bnfmore{}{\bnfpn{sec\_icl\_examples}, \bnfpn{sec\_knowledge},\bnfpn{sec\_format}, \bnfpn{sec\_input}, \bnfpn{sec\_context}, \bnfpn{sec\_cot} }\\
\bnfprod{X\_persona\_prompt}{\bnfts{"You are an expert labeller for labelling the overall sentiment}}\\ \bnfmore{}{\bnfts{conveyed in a product review. You may need to consider expert-level}}\\ \bnfmore{}{\bnfts{knowledge. Where required, consider what the lay person would think."}}\\
 \bnfprod{X\_task\_prompt}{\bnfts{"Your task is to classify the sentence \_\_TASK\_INPUT\_0\_\_ by its sentiment."}}\\
 \bnfprod{X\_icl\_prompt}{\bnfts{"Here are some examples based on unrelated scenarios:"}}\\
  \bnfprod{X\_format\_prompt}{\bnfts{"Conclude your final answer in the following format:}}\\ \bnfmore{}{\bnfts{\{'Thought': 'Your step by step analysis', 'Answer': 'Your final answer'\}"}}\\
  \bnfprod{X\_context\_prompt}{\bnfts{"To perform this task, scrutinise the following context: \_\_CONTEXT\_\_"}}\\
  \bnfprod{X\_cot\_prompt}{\bnfts{"Before providing your response, thoroughly examine the context, think}}\\ \bnfmore{}{\bnfts{methodically, and then present your answer in the required format."}}\\
  \bnfprod{X\_icl\_examples}{\bnfts{"['\_\_ICL\_0\_\_','\_\_ICL\_1\_\_','\_\_ICL\_2\_\_','\_\_ICL\_3\_\_','\_\_ICL\_4\_\_']"}}\\
\end{bnf*}

\begin{multicols}{2}
\textbf{Persona Section}
\begin{bnf*}
\bnfprod{sec\_persona}{\bnfpn{persona\_expr} }\\
\bnfmore{}{\bnfor \bnfpn{null\_string}}\\
\bnfprod{persona\_expr}{\bnfpn{text\_op} \bnfpn{persona\_expr} ) }\\
\bnfmore{}{ \bnfor \bnfpn{X\_persona\_prompt}}\\
\end{bnf*}
\textbf{ In-Context Learning Section }
\begin{bnf*}
\bnfprod{sec\_icl}{\bnfpn{icl\_expr} \bnfts{+} \bnfpn{icl\_examples}}\\
\bnfmore{}{ \bnfor \bnfpn{null\_string}}\\
\bnfprod{icl\_expr}{\bnfpn{text\_op}\bnfpn{icl\_expr} ) }\\
\bnfmore{}{\bnfor \bnfpn{X\_icl\_prompt}}\\
\bnfprod{icl\_examples}{\bnfpn{edit\_op}\bnfpn{icl\_examples} ) }\\
\bnfmore{}{\bnfor \bnfpn{X\_icl\_examples}}\\
\end{bnf*}
\textbf{ Task Section }
\begin{bnf*}
\bnfprod{task\_persona}{\bnfpn{task\_expr} }\\
\bnfprod{task\_expr}{\bnfpn{text\_op}\bnfpn{task\_expr} }\\
\bnfmore{}{\bnfor \bnfpn{X\_task\_prompt}}\\
\end{bnf*}

\columnbreak
\textbf{ Format Section }
\begin{bnf*}
\bnfprod{sec\_format}{\bnfpn{format\_expr} \bnfts{+} \bnfpn{format\_examples}}\\
\bnfmore{}{\bnfor \bnfpn{format\_expr}}\\
\bnfmore{}{\bnfor \bnfpn{null\_string}}\\
\bnfprod{format\_expr}{\bnfpn{text\_op}\bnfpn{format\_expr} }\\
\bnfmore{}{\bnfor \bnfpn{X\_format\_prompt}}\\
\end{bnf*}

\textbf{ Context Section }
\begin{bnf*}
\bnfprod{sec\_context}{\bnfpn{context\_expr} \bnfts{+} \bnfpn{context\_examples}}\\
\bnfmore{}{\bnfor \bnfpn{context\_expr}}\\
\bnfmore{}{\bnfor \bnfpn{null\_string}}\\
\bnfprod{context\_expr}{\bnfpn{text\_op}\bnfpn{context\_expr} }\\
\bnfmore{}{\bnfor \bnfpn{X\_context\_prompt}}\\
\end{bnf*}

\textbf{ CoT Section }
\begin{bnf*}
\bnfprod{sec\_cot}{\bnfpn{cot\_expr}}\\
\bnfmore{}{\bnfor \bnfpn{null\_string}} \\
\bnfprod{cot\_expr}{\bnfpn{text\_op>} \bnfpn{cot\_expr>} )}\\
\bnfmore{}{\bnfor\bnfpn{X\_cot\_prompt}} \\
\end{bnf*}

\end{multicols}

\textbf{ Edit Operators }
\begin{bnf*}
\bnfprod{text\_op}{\bnfpn{edit\_op} \bnfor \bnfpn{dict\_transformation\_op} \bnfor \bnfpn{llm\_transformation\_op}} \\
\bnfprod{edit\_op}{swap\_elements(index1=\bnfpn{view\_index\_expr},index2=\bnfpn{view\_index\_expr},level=\bnfpn{chunk\_level},texts=} \\
\bnfmore{\bnfor remove\_element(index=\bnfpn{view\_index\_expr},level=\bnfpn{chunk\_level},texts=} \\
\bnfmore{\bnfor readd\_element(index=\bnfpn{atomic\_view\_index\_expr},level=\bnfpn{chunk\_level},texts=} \\
\bnfmore{\bnfor duplicate\_element(index1=\bnfpn{view\_index\_expr},index2=\bnfpn{atomic\_view\_index\_expr},level=\bnfpn{chunk\_level},texts=} \\
\bnfprod{dict\_transformation\_op}{remove\_stopwords(index=\bnfpn{view\_index\_expr},texts=} \\
\bnfmore{\bnfor synonimise(index=\bnfpn{view\_index\_expr},texts=} \\
\bnfprod{llm\_transformation\_op}{paraphrase(index=\bnfpn{view\_index\_expr},texts=} \\
\bnfmore{\bnfor summarise(percent=\bnfpn{tenths},index=\bnfpn{view\_index\_expr},texts=} \\
\bnfprod{chunk\_level}{sentence \bnfor phrase \bnfor word} \\
\bnfprod{view\_index\_expr}{\bnfpn{atomic\_view\_index\_expr} \bnfor \bnfpn{slice\_view\_index\_expr}} \\
\bnfprod{atomic\_view\_index\_expr}{[\bnfpn{zero\_to\_nine,}]} \\
\bnfprod{slice\_view\_index\_expr}{[\bnfpn{zero\_to\_nine},\bnfpn{zero\_to\_nine}]} \\
\bnfprod{null\_string}{\bnfts{" "}} \\
\bnfprod{non\_zero\_digit}{1 \bnfor 2 \bnfor 3 \bnfor 4 \bnfor 5 \bnfor 6 \bnfor 7 \bnfor 8 \bnfor 9} \\
\bnfprod{tenths}{0.\bnfpn{non\_zero\_digit}} \\
\bnfprod{zero\_to\_nine}{\bnfpn{non\_zero\_digit} \bnfor 0}
\end{bnf*}

\newpage
\section{Grammar-guided GP Settings and Hyperparameters}\label{app:settings}
\begin{table}[!h]
\caption{Grammar-guided GP Settings and Hyperparameters}
\label{tab:settings}
\centering
\begin{tabular}{|l|l|}
\hline
\textbf{Parameter} & \textbf{Value} \\
\hline
Evaluate N individuals on Validation Set per Generation & 1 \\
Genetic Programming System (GP) & Context-Free Grammar Genetic Programming on Serialized Trees \\
GP max Nodes & 1024 \\
Maximum ICL Available & 5 \\
Offspring Population Size & 50 \\
Parent Selection Operator & Tournament \\
Parent Tournament Selection Size & 2 \\
Population Initialisation Method & Probabilistic Tree Creation 2 \cite{luke2000two} \\
Population Size & 50 \\
Resample Training Data every & Generation \\
Rows of Training Data & 20 \\
Search Implementation & Alogos\footnote{https://github.com/robert-haas/alogos}  \\
Survivor Selection Operator & Tournament \\
Survivor Tournament Selection Size & 4 \\
Number of Generations & 20 \\
\hline
\end{tabular}
\end{table}

\section{Surrogate Model Settings and Hyperparameters}

\begin{table}[h]
    \centering
\caption{Surrogate Model Parameters}
    
    \begin{tabular}{|l|l|}
        \hline
                Parameter & Value \\
        \hline
        \multicolumn{2}{|c|}{\textbf{Tunable Parameters}} \\
        \hline

        Submodel Layer Sizes & $[(128,1) , (128,64,1) , (128,64,32,1)]$ \\
        Dropout Rate & [0.0 , 0.1 , 0.2 , 0.5] \\
        Batch Size & [16 , 32] \\
        Learning Rate & $[10^{-4} , 10^{-3}]$ \\
        \hline
        \multicolumn{2}{|c|}{\textbf{Non-Tunable Parameters}} \\
        \hline

        Activation Function & tanh \\
        Input & Embeddings of Prompt Template \\
        Input Encoder & \texttt{sentence-transformers/all-MiniLM-L6-v2} \\
        Epochs & 200 \\
        Submodel Count & 10 \\
        Train-Val Split & 70:30 \\
        Hyperparameter Cross-Validation Folds & 5 \\
        Combinations of Tunable Hyperparameters Tested & 10\\ 
        Model Selection Criterion & Minimum Validation Loss averaged across submodels\\
        Loss & Mean Squared Error \\
        Optimiser & Adam\\
        \hline
    \end{tabular}
\end{table}

\section{LLM Settings}\label{app:llmsettings}
\begin{table}[h!]
\caption{LLM Parameters}
\label{tab:llmargs}
\centering
\begin{tabular}{|l|l|}
\hline
\textbf{Parameter} & \textbf{Value} \\
\hline
Max New Tokens & 2048 \\
Sampling & False \\
Temperature & 0.0 \\
Quantisation & None \\
Source for Pretrained Weights & \url{https://huggingface.co/}\\
HuggingFace Model IDs: &\\
\hphantom{00} -- Llama3.2 3B & \texttt{meta-llama/Llama-3.2-3B-Instruct}\\
\hphantom{00} -- Llama3 8B & \texttt{meta-llama/Meta-Llama-3-8B-Instruct}\\
\hphantom{00} -- Gemma2 9B & \texttt{google/gemma-2-9b-it}\\
\hphantom{00} -- Qwen 2.5 32B & \texttt{Qwen/Qwen2.5-32B-Instruct}\\
\hline
\end{tabular}
\end{table}

\newpage
\section{Training-Set Fitness Curves}\label{app:learningcurves}

\begin{figure}[!h]
    \centering
    \includegraphics[width=\textwidth]{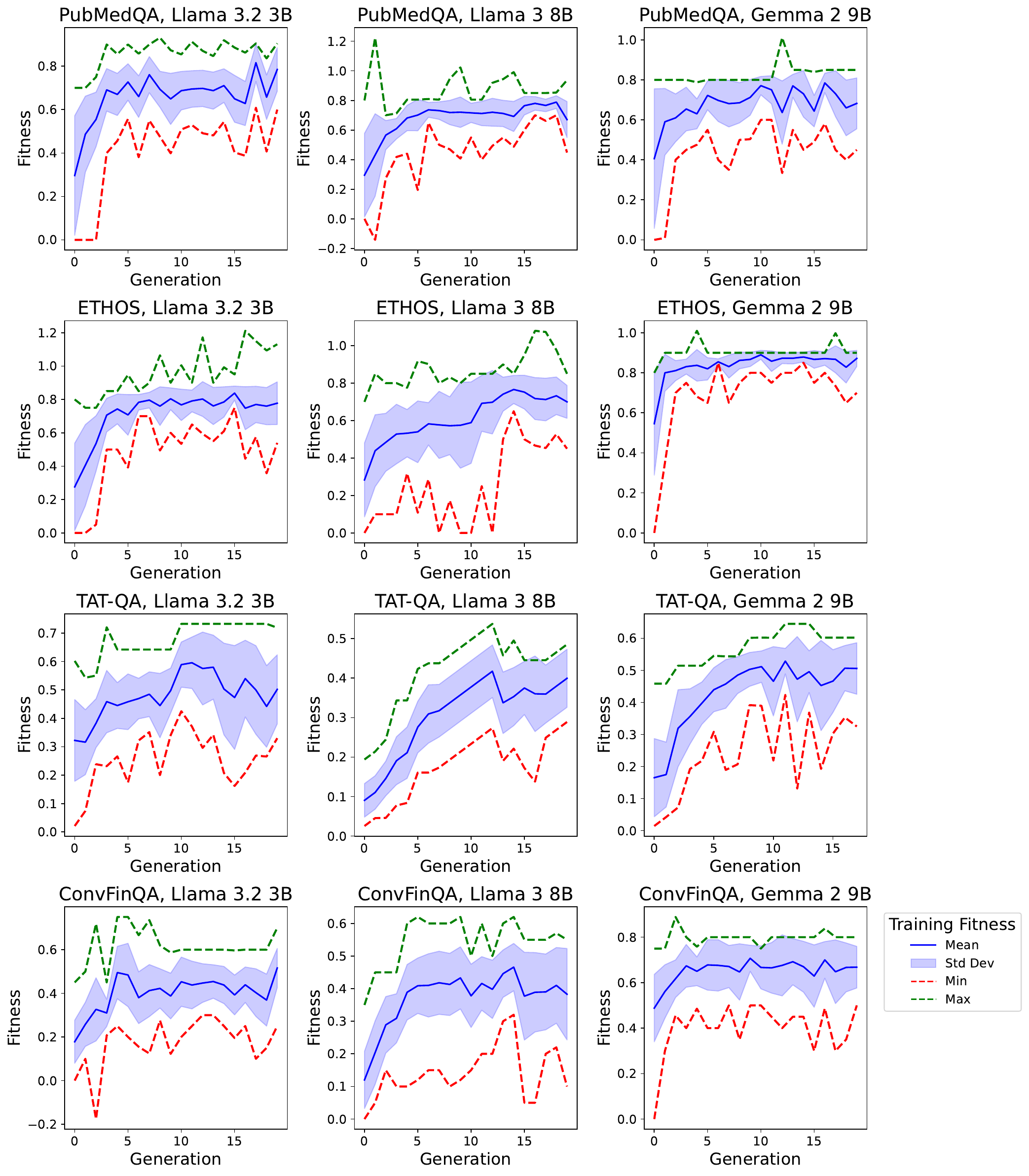}
    \caption{Mean and Standard Deviation of the Training Fitness at every generation for our Grammar-guided Search.}

\end{figure}

\newpage
\section{Baseline Prompt Templates}\label{app:base-templates}

\subsection{PubMedQA}
This prompt was designed based on the prompt used in \cite{pubmedtemplate}.
\begin{lstlisting}[style=promptstyle]
"""
You are a Lead Physician responsible for answering challenging medical questions.
Carefully analyse the Research Paper Abstract given in [Context] and identify keypoints relevant to the [Question]. Then, answer the question in [Question] according to the [Task Constraints] and [Output Constraints].
You provide clear, concise, evidence-based answers and focus on precision while considering research nuances. 
## [Task Constraints]
Answer the question in the [Instruction] according to the [Output Constraints], which is in JSON format: {'Thought':'...', 'Answer':'...'}. In the 'Thought' section, provide the complete thought process, and in the 'Answer' section, provide a conclusion that fully meets the requirements of the [Instruction].\nIn the thought process, combine the key points from the [Question] and [Context] to analyse the data, and provide detailed reasoning.
The [Context] is used to provide background information for the task. If you need to extract information from it, you can use it. If it is empty, ignore it.
Simplify the answer and ensure that the 'Answer' is provided.
## [Output Constraints]
{'Thought': 'A thorough explanation or a step by step analysis of the [Context] answer [Question] ', 'Answer': 'yes/no/maybe'}.
Think carefully step by step and apply the above information to the [Question] and generate the **Answer** following the specificed [Task Constraints]
"""
\end{lstlisting}
\subsection{ETHOS-Binary}
This prompt was designed based on the prompt used in \cite{fernando2023apopromptbreeder}.
\begin{lstlisting}[style=promptstyle]
"""
You are an expert content moderator for a social media site.
Carefully analyse the following [Input] whether it constitutes Hate Speech.
You should provide clear, concise, evidence-based answers and focus on precision while considering social nuances. In the 'Thought' section, provide the complete thought process, and in the 'Answer' section, provide a conclusion that fully meets the requirements of the [Instruction]. In the thought process, combine the key points from the <input data>, [Instruction], [Knowledge] to analyse the data, and provide detailed reasoning. The [Knowledge] and <Hints> are golden experiences used to analyse the <input data>. You can refer to them to analyse the data. If they are empty, ignore them. If the [Input] is too long, simplify the answer and ensure that the 'Answer' is provided.
## [Knowledge]
Text contains hate speech if it includes any of the following: Words or phrases that are derogatory, disrespectful, or dehumanizing toward a particular group of people. Statements that generalize about a particular group of people in a negative way. Statements that incite hatred or violence against a particular group of people. Statements that express a desire for the extermination or removal of a particular group of people. Text contains hate speech if it contains language that is hostile or discriminatory towards a particular group of people. language that is racist, sexist, homophobic, or other forms of hate speech. Hate speech is harmful because it can create a hostile environment for marginalized groups and can lead to discrimination and violence.
## [Instruction]
analyse the following comment and determine if it is 'Hateful' or 'Not Hateful'.
## [Output Constraints]
Respond in strict JSON format with a key-value pairs: {'Thought': 'A thorough explanation or a step by step analysis of the [Context] answer [Question] ', 'Answer': 'Hateful' | 'Not Hateful'}.
Think carefully step by step and apply the above information to the [Question] and generate the **Answer** following the specificed [Task Constraints].
"""
\end{lstlisting}
\subsection{TAT-QA}
This prompt was designed based on the prompt used in \cite{tattemplate}.
\begin{lstlisting}[style=promptstyle]
"""
You are a data analysis with expertise in interpreting tabular data.
Please analyse the [Table] based on the [Paragraphs] , and answer the question in the [Instruction] according to the [Task Constraints] and [Output Constraints].
Below is an instruction that describes a question answering task in the finance domain, paired with an input table and its relevant text that provide further context.The given question is relevant to the table and text.
## [Task Constraints]
Given a table and a list of texts in the following, what is the answer to the question? You may need to apply these steps: 1.Extract the relevant numerical values from the provided table or texts as 'evidence'. 2. IGenerate an equation using the extracted numerical values as 'equation'. 3. Calculate the answer based on the equation for the final 'answer'.Answer the question in the [Instruction] according to the [Output Constraints], which is in JSON format: {'Answer':'...'}. In the 'Answer' section, provide a conclusion that fully meets the requirements of the [Instruction].In the thought process, combine the key points from the <input data>, [Instruction], [Knowledge], and <Hints> to analyse the data, and provide detailed reasoning.The [Context] is used to provide background information for the task. If you need to extract information from it, you can use it. If it is empty, ignore it.The [Knowledge] and <Hints> are golden experiences used to analyse the <input data>. You can refer to them to analyse the data. If they are empty, ignore them.If the [Input] is too long, simplify the answer and ensure that the 'Answer' is provided.
## [Output Constraints]
Finally, present the calculated answer in the format: {'Answer': 'Your Answer'}. Before answering, you must judge whether you can obtain information directly related to [Instruction] from [Experience Knowledge] and [Context]. You only copy the parts directly related to [Instruction]. Please think carefully step by step and apply the above information to the [Instruction] and generate the **answer** following the specificed [Task Constraints]
"""
\end{lstlisting}
\subsection{ConvFinQa}
This prompt was designed based on the prompt used in \cite{chen2022convfinqa}.
\begin{lstlisting}[style=promptstyle]
"""
You are a financial analyst with expertise in interpreting tabular data and conversations.
Please analyse the [Table] based on the [Text] , and answer propose a program that would help answer the question in the [Instruction] according to the [Task Constraints] and [Output Constraints].
Below is an instruction that describes a question answering task in the finance domain, paired with an input table and its relevant text that provide further context. The given question is relevant to the table and text. 
## [Task Constraints]
Given a table and texts above and below it in the following, how could you answer the question? You need to craft a short numerical program that computes the answer. For this program, you can use the following functions: add(n1,n2), subtract(n1,n2), multiply(n1,n2), divide(n1,n2), exp(n1,n2), greater(n1,n2), table_max, table_min, table_sum and table_average. table_average() Returns the average of all values in the table. table_sum() Returns the sum of all values in the table. table_min() Returns the minimum value in the table. table_max() Returns the maximum value in the table. 
1.Extract the relevant numerical values from the provided table or texts. 2. Generate an equation using the extracted numerical values as 'Formula'. 3. Calculate the answer based on the equation for the final 'Answer'. Answer the question in the [Instruction] according to the [Output Constraints], which is in JSON format: {'Formula':'...','Answer':'...'}. In the 'Formula' section, provide a numerical answer with decimal point or with %; that fully meets the requirements of the [Instruction]. In the thought process, combine the key points from the <input data>, [Instruction], [Knowledge], and <Hints> to analyse the data, and provide detailed reasoning. The [Context] is used to provide background information for the task. If you need to extract information from it, you can use it. If it is empty, ignore it. The [Knowledge] and <Hints> are golden experiences used to analyse the <input data>. You can refer to them to analyse the data. If they are empty, ignore them. If the [Input] is too long, simplify the answer and ensure that the 'Formula' is provided.
## [Output Constraints]
Finally, present the calculated answer in the format: {'Formula': 'Your Program','Answer': 'Your final numerical answer'}. Before answering, you must judge whether you can obtain information directly related to [Instruction] from [Experience Knowledge] and [Context]. You only copy the parts directly related to [Instruction].
Please think carefully step by step and apply the above information to the [Instruction] and generate the  'Formula' and 'Answer'following the specificed [Task Constraints]
"""
\end{lstlisting}
\section{Prompt Template for LLM-based Edits}
\subsection{Paraphrase}
\begin{lstlisting}[style=promptstyle]
"""Paraphrase the following text while maintaining as much of the original meaning as possible. Give the paraphrased answer in JSON format as follows: {{"answer": "your paraphased text"}}. 

The original text: 
```
{input_text}
```"""
\end{lstlisting}
\subsection{Summarise}
\begin{lstlisting}[style=promptstyle]
"""Reduce the text length of this text by slightly rephrasing and give the final answer in JSON format as follows: {{"answer": "your shortened text"}}. The length of your output should be approximately {ratio}\% of the length of the original text.

The original_text: 
```
{input_text}
```"""
\end{lstlisting}

\end{document}